\documentclass[lettersize,journal]{IEEEtran}
\usepackage{amsmath,amsfonts}
\usepackage{algorithmic}
\usepackage{algorithm}
\usepackage{array}
\usepackage[caption=false,font=normalsize,labelfont=sf,textfont=sf]{subfig}
\usepackage{textcomp}
\usepackage{stfloats}
\usepackage{url}
\usepackage{verbatim}
\usepackage{graphicx}
\usepackage{cite}
\usepackage{xcolor}
\usepackage{multicol}
\usepackage{multirow}
\usepackage{booktabs} 

\usepackage{hyperref}

\def\eg{\emph{e.g.}}

\def\ie{\emph{i.e.}}

\begin{document}

\title{NMS: Efficient Edge DNN Training via \underline{N}ear-\underline{M}emory \underline{S}ampling on 
 Manifolds}


 \author{Boran Zhao$^{1,\star}$, Haiduo Huang$^{2,\star}$, Qiwei Dang$^{2}$, Wenzhe Zhao$^{2}$, Tian Xia$^{2}$~\IEEEmembership{Member,~IEEE},\\
        and Pengju Ren$^{2}$~\IEEEmembership{Member,~IEEE}
        \thanks{- Boran Zhao and Haiduo Huang contributed equally to this work.}
        \thanks{- $^{1}$Boran Zhao is with the School of Software Engineering, Xi'an Jiaotong University, Xi'an, Shaanxi, China.}
        \thanks{- $^{2}$All other authors are with the National Key Laboratory of Human-Machine Hybrid Augmented Intelligence, National Engineering Research Center of Visual Information and Applications, and Institute of Artificial Intelligence and Robotics, Xi'an Jiaotong University, Xi’an, Shaanxi, China.}
        \thanks{- E-mail: pengjuren@xjtu.edu.cn (Corresponding Author).}
    }

\markboth{Journal of \LaTeX\ Class Files,~Vol.~14, No.~8, August~2021}%
{Shell \MakeLowercase{\textit{et al.}}: A Sample Article Using IEEEtran.cls for IEEE Journals}

\maketitle

\begin{abstract}
Training deep neural networks (DNNs) on edge devices has attracted increasing attention due to its potential to address challenges related to domain adaptation and privacy preservation. However, DNNs typically rely on large datasets for training, which results in substantial energy consumption, making the training in edge devices impractical.
Some dataset compression methods have been proposed to solve this challenge. For instance, the coreset selection and dataset distillation reduce the training cost by selecting and generating representative samples respectively. Nevertheless, these methods have two significant defects: (1) The necessary of leveraging a DNN model to evaluate the quality of representative samples, which inevitably introduces inductive bias of DNN, resulting in a severe \textit{generalization issue}; (2) All training images require multiple accesses to the DDR via long-distance PCB connections, leading to substantial energy overhead. To address these issues, inspired by the nonlinear manifold stationary of the human brain, we firstly propose a \textit{DNN-free} sample-selecting algorithm, called DE-SNE, to improve the generalization issue. Secondly, we innovatively utilize the near-memory computing technique to implement DE-SNE, thus only a small fraction of images need to access the DDR via long-distance PCB. It significantly reduces DDR energy consumption. As a result, we build a novel expedited DNN training system with a more efficient in-place Near-Memory Sampling characteristic for edge devices, dubbed NMS. As far as we know, our NMS is the first \textit{DNN-free} near-memory sampling technique that can effectively alleviate generalization issues and significantly reduce DDR energy caused by dataset access. The experimental results show that our NMS outperforms the current state-of-the-art (SOTA) approaches, namely DQ, DQAS, and NeSSA, in model accuracy. For example, our method achieves average accuracy improvements of 11.9\% over DQ, 9.7\% over DQAS, and 4.7\% over NeSSA on the ImageNet-1K dataset. The maximum improvements are 24.7\%, 24.3\%, and 14.4\%, respectively. Moreover, in terms of hardware efficiency, our NMS reduces DDR energy consumption by an average of 5.3 times and 50.4 times compared to NeSSA and DQ, respectively. The overall system efficiency is improved by 1.2 times compared to the sparse training accelerator, \ie, THETA.

\end{abstract}

\section{Introduction}
\begin{figure}[!t]
\centering
\includegraphics[width=1.0\columnwidth]{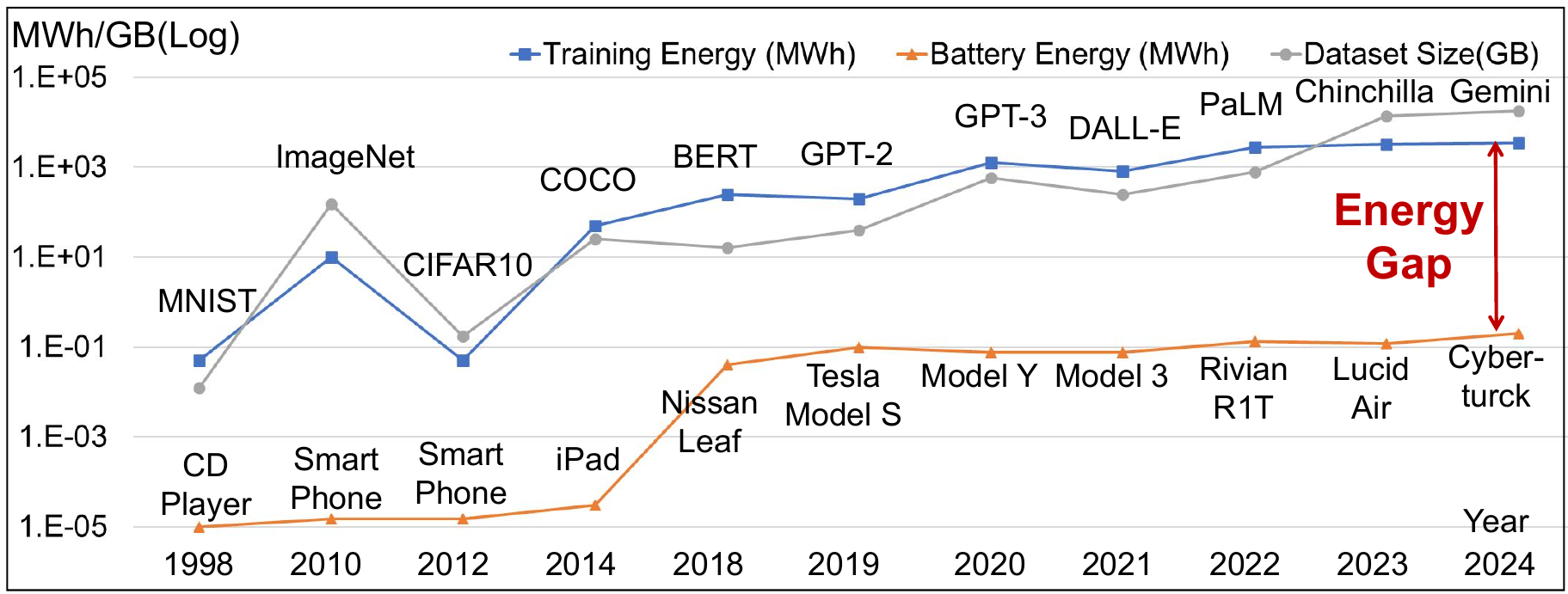} 
\caption{The gap between the energy required for training on edge devices and their battery capacity.}
\label{fig_train_battery_energy} 
\end{figure}

\IEEEPARstart{D}{eep} Neural Networks (DNNs) have achieved remarkable success in various fields \cite{deep_inst0,deep_inst1,vit,shufflenetv2,pow2,remap,hipu}. Current DNNs typically rely on large-scale datasets to enhance model accuracy by training in the cloud servers, which results in considerable energy consumption. As depicted in Fig. \ref{fig_train_battery_energy} (blue line), training a DNN nowadays requires approximately $\mathcal{O}(10^{3})$ MWh, which is enough to power 94 households for an entire year. In addition, due to the domain mismatch between training data and real-world data, retraining or fine-tuning is often required when adapting a DNN to a new scenario. The training cost is very large. Moreover, the way of training DNNs in cloud servers raises data privacy concerns and communication latency issues. Thus, there is an increasing need to train DNNs on edge devices. Nevertheless, the edge devices are typically battery-powered, which makes it challenging to handle the large energy consumption associated with training on large-scale datasets, as shown in Fig. \ref{fig_train_battery_energy} (orange and gray lines). This raises two critical questions: Is all the data in a large-scale dataset necessary? If redundant data is present, can it be eliminated to reduce edge training energy without degrading the accuracy of the DNN? This paper primarily aims to tackle these two pivotal issues.

Several dataset distillation (DD) methods \cite{dd_inst0,dd_inst1,dd_inst2,dd_inst3,dm} generate representative synthetic samples by optimizing images as trainable parameters. However, DD needs an extra outer loop in the normal training process, which results in severe energy costs. The training cost is usually several orders of magnitude higher than traditional DNN training. To address this challenge, some works have proposed efficient coreset methods aimed at selecting a small subset of representative samples. Although this method reduces training costs compared to the DD methods, it increases the computational cost due to requiring the results of DNN forward pass as features and involving multiple iterations to extract samples. Additionally, both DD and coreset methods rely on specific DNN models to evaluate the differences between representative samples and real samples, which results in a severe generalization issue. For instance, the SOTA work DQ\cite{dq} will drop accuracy up to 52.5\% when switching DNN on the CIFAR10 dataset. Therefore, how to compress datasets while maintaining their generalization ability remains an unresolved challenge.

\begin{figure}[ht]
\centering
\includegraphics[width=1\columnwidth]{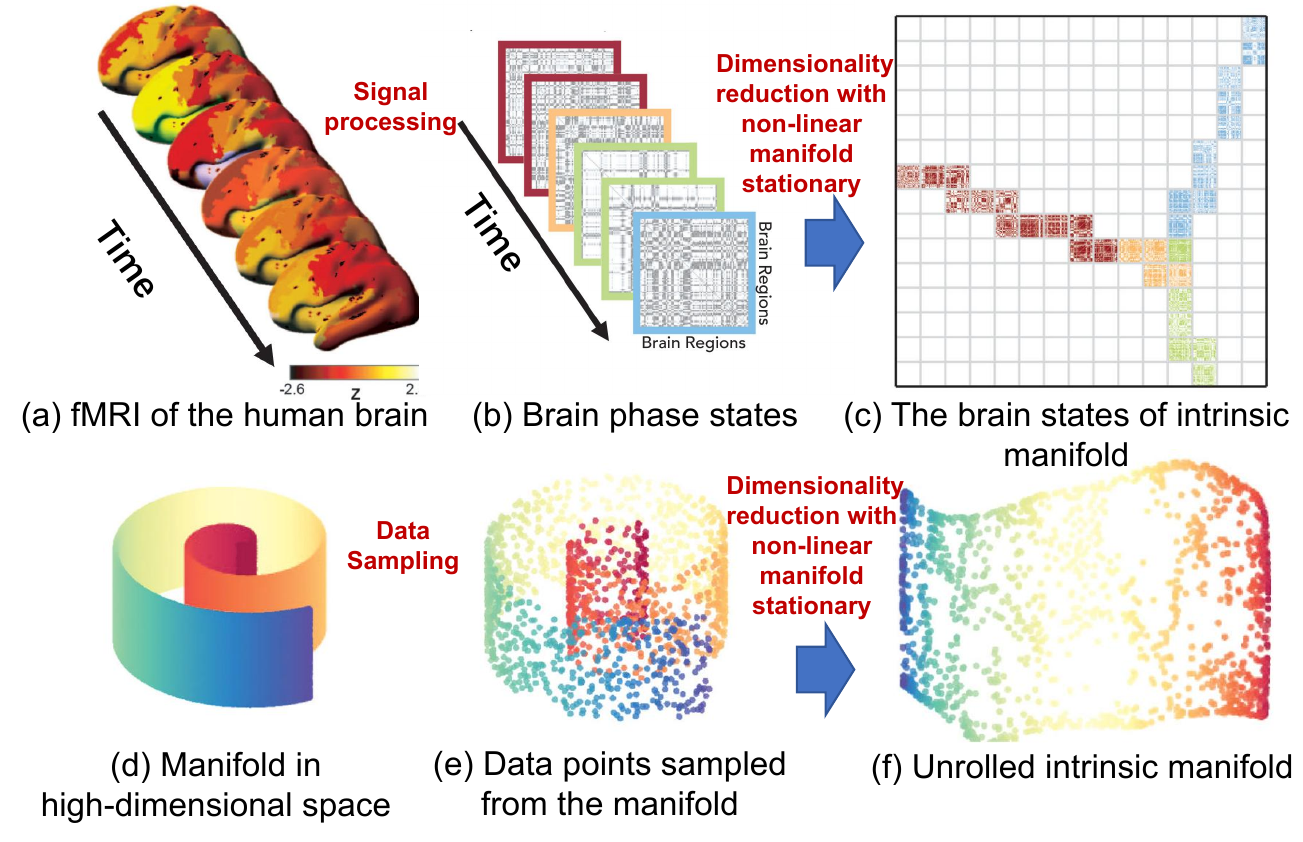} 
\caption{The human brain processes information through \textit{non-linear manifold stationary}.\cite{gao2021nonlinear} The data sampling imitates the brain information processing.} 
\label{fig_brain_manifold}
\end{figure}

To tackle these challenges, we introduce a brain-inspired {\bf N}ear-{\bf M}emory {\bf S}ampling method called {\bf NMS}, designed to enhance training efficiency on edge devices. Our NMS is based on two key insights: (1) The human brain possesses more efficient learning and expressive capabilities compared to DNNs. Research~\cite{gao2021nonlinear} indicates that the human brain, through long-term natural evolution, processes information with low-dimensional manifold stationary. This means that it can map the correlations between high-dimensional image samples into a low-dimensional manifold space using nonlinear manifold transformations, as shown in Fig.~\ref{fig_brain_manifold}. By emulating this low-dimensional manifold stationary, our method aims to significantly enhance both sampling efficiency and sample generalizability. (2) The redundancy inherent in large-scale datasets results in substantial energy consumption due to excessive data movement. For example, the energy consumption from DRAM data movement can exceed 40\% \cite{dq} in ImageNet-1K training. To address this, we first propose an \textit{DNN-free} dataset sampling algorithm, called DE-SNE. It overcomes traditional t-SNE unstable problems caused by perplexity search. Second, we use a near-memory computing architecture to implement DE-SNE within DRAM logic circuits. 

Our main contributions can be summarized as follows:

\begin{itemize}
\item Drawing inspiration from the human brain's low-dimensional manifold stationary in processing image information, we are the first to introduce a simple but efficient low-dimensional manifold space sampling algorithm based on t-SNE. This approach ensures uniform sample distribution and enhances dataset generalizability while eliminating the overhead of multiple DNN forward inferences typically required in traditional sampling methods.

\item The computation bottleneck of the t-SNE algorithm is analyzed. We identify that the binary search process can increase perplexity errors and potentially render the t-SNE algorithm unstable. A differential evolution algorithm is proposed to replace binary search, called DE-SNE, effectively reducing perplexity search errors within acceptable computational costs.

\item Developing a manifold space sampling circuit based on near-memory computing architecture can significantly mitigate the ``energy wall" issue between DRAM and accelerators during model training. This reduction in energy consumption makes it feasible to train DNNs on edge devices.
\end{itemize}

\section{Background and Motivation}
\subsection{Information Processing in the Human Brain}

\begin{figure}[ht]
\centering
\includegraphics[width=0.9\columnwidth]{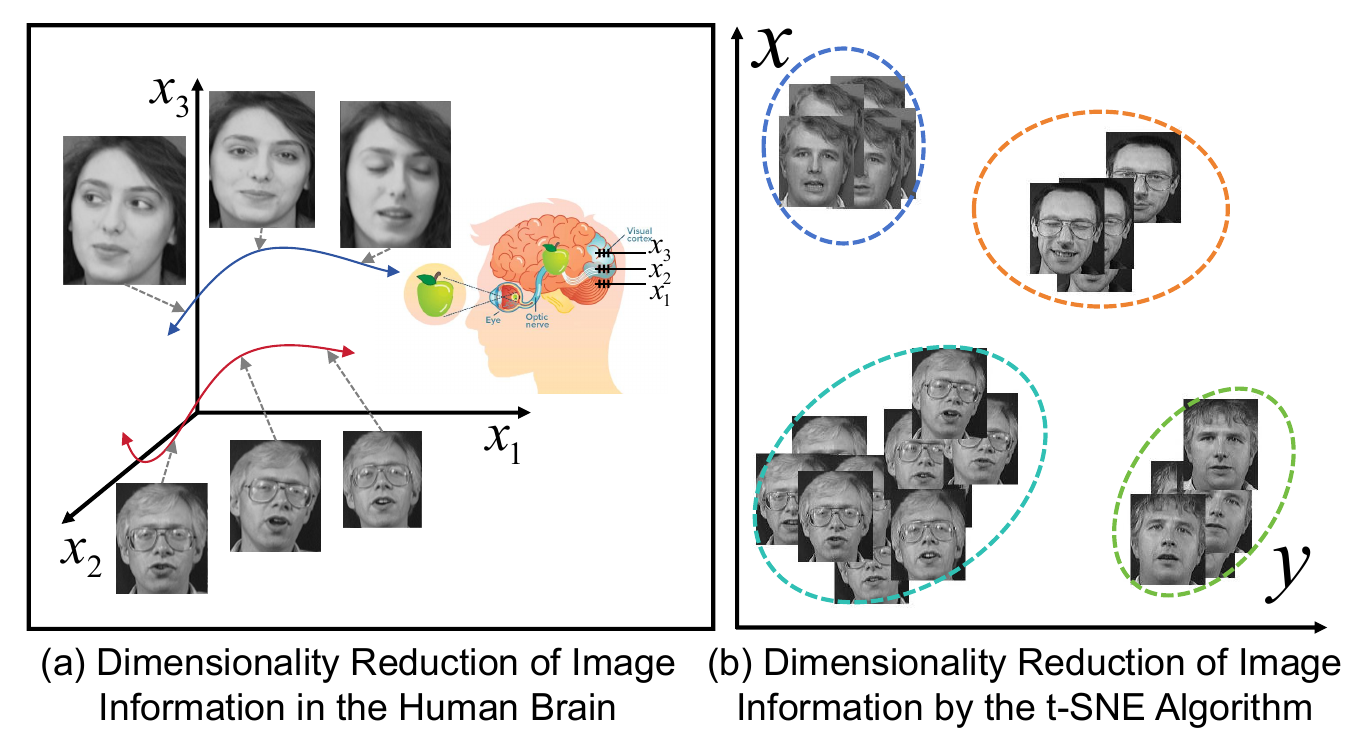} 
\caption{Both the human brain (a) \cite{manifold0} and the t-SNE method (b) transform high-dimensional images into a low-dimensional manifold space while preserving the correlation (and distance) between images, a process referred to as \textit{non-linear manifold stationarity}.}
\label{fig_face_brain_tsne} 
\end{figure}

A manifold refers to a space with local Euclidean properties, a concept derived from differential geometry. As shown in Fig.~\ref{fig_face_brain_tsne} (a), the research teams led by Professor H. Sebastian Seung from MIT \cite{manifold0}, Professor Sam T. Roweis from the University of London \cite{manifold1}, and Professor Joshua B. Tenenbaum from Stanford University \cite{manifold2} independently publish papers in Science, proposing that the brain exhibits nonlinear manifold stationary when processing visual information. This stability allows for the transformation of different high-dimensional image information into a low-dimensional manifold space while preserving the similarity between images in the high-dimensional space. Furthermore, research teams led by Professor Gustavo Deco from Pompeu Fabra University \cite{rue2021decoding} and Professor Dustin Scheinost from Yale University \cite{gao2021nonlinear} experimentally demonstrated the low-dimensional manifold stationary of brain dynamics using functional magnetic resonance imaging (fMRI) and electroencephalography (EEG). These findings in neuroscience offer heuristic insights into efficiently extracting image samples during DNN training.

\subsection{Nonlinear Manifold Dimensionality Reduction Algorithms}
Nonlinear manifold dimensionality reduction algorithms are inspired by the low-dimensional manifold stationary observed in the human brain. One such algorithm is t-SNE \cite{tsne}, proposed by Hinton Geoffrey, which is widely used for high-dimensional data. It excels at preserving the local structure of the data, ensuring that similar points remain neighbors after dimensionality reduction. As depicted in Figure~\ref{fig_face_brain_tsne} (b), the t-SNE algorithm maintains the approximate image features of the same person while ensuring that the features of different individuals are separated. Thus, t-SNE theoretically can be employed to develop an efficient image sample extraction method analogous to the human brain's processes. The specific steps of the t-SNE algorithm are as follows:

\subsubsection{Constructing Probability Distributions in High-Dimensional Feature Space to Measure Sample Similarity}

Given a dataset {\small \( X \)} consisting of {\small \( N \)} samples, let \( x_i \) and \( x_j \) be two arbitrary sample points. These points follow Gaussian distributions centered at \( x_i \) and \( x_j \) with variances \( \sigma_i^2 \) and \( \sigma_j^2 \), respectively. The conditional probability of similarity between \( x_i \) and \( x_j \) is defined as:
\begin{equation}\label{math_pji}  \small
p_{j|i} = \frac{\exp\left(-\|x_i - x_j\|^2 / 2\sigma_i^2\right)}{\sum_{k \neq i}\exp\left(-\|x_i - x_k\|^2 / 2\sigma_i^2\right)}
\end{equation}

The joint similarity between \( x_i \) and \( x_j \) is:
\begin{equation} \small
p_{ij} = \frac{p_{j|i} + p_{i|j}}{2N}
\end{equation}

\subsubsection{Determining \( \sigma_i \) via Binary Search for Perplexity}

The perplexity \( P_i \) of the distribution is defined by the entropy \( H_i \):
\begin{equation}\label{per_sigma1} \small
P_i = 2^{H_i}, \; 
H_i = -\sum_{j} p_{j|i} \log_2 p_{j|i}
\end{equation}

\subsubsection{Constructing Probability Distributions in Low-Dimensional Manifold Space}

Let {\small \( Y \)} represent the low-dimensional embeddings of the high-dimensional dataset {\small \( X \)}. In this low-dimensional space, data points follow a Student's t-distribution with one degree of freedom. The joint probability \( q_{ij} \) between points \( y_i \) and \( y_j \) in the low-dimensional manifold space is defined as:
\begin{equation}\label{math_qij} \small
q_{ij} = \frac{(1 + \|y_i - y_j\|^2)^{-1}}{\sum_{k \neq l} (1 + \|y_k - y_l\|^2)^{-1}}
\end{equation}

\subsubsection{Computing Low-Dimensional Manifold Space Coordinates}

The similarity between the high-dimensional feature space and the low-dimensional manifold space probability distributions is measured using the Kullback-Leibler (KL) divergence:
\begin{equation} \small
KL(P\|Q) = \sum_{i \neq j} p_{ij} \log \left(\frac{p_{ij}}{q_{ij}}\right)
\end{equation}

To find the sample embedding coordinates in the low-dimensional manifold space, the KL divergence is minimized using gradient descent. The gradient is computed as follows:
\begin{equation} \small
\frac{\partial KL}{\partial y_i} = 4 \sum_{j} (p_{ij} - q_{ij})(y_i - y_j)(1 + \|y_i - y_j\|^2)^{-1}
\end{equation}

\section{Related Work}

\subsection{Dataset Distillation}

Dataset Distillation (DD) is formally introduced by Wang et al.~\cite{wang2018dataset}. This technique generates a synthetic dataset using the principle of minimizing loss, ensuring that the synthetic dataset matches the accuracy of the real dataset. Dataset distillation methods can be divided into meta-learning methods and data-matching methods. The former sets the synthetic data to be distilled as learned parameters and optimizes them in a nested loop manner based on model training risk~\cite{deng2022remember}. The latter distills synthetic data by mimicking the impact of real data on parameters or feature spaces of models, which mainly includes parameter matching, gradient matching, and trajectory matching. The parameter matching primarily refers to distribution matching~\cite{dm}. However, the gradient matching~\cite{zhao2020dataset,kim2022dataset} is a short-range parameter matching method that requires the calculation of second-order gradients, resulting in high complexity. To address this issue, researchers have proposed trajectory matching~\cite{dd_inst1,liu2023dream}, which avoids the computation of second-order gradients but introduces a recursive computation graph. In contrast, our NMS only exploits a simple and efficient DE-SNE dimension-reduction function for real images without a complex and time-consuming dataset-matching process while avoiding the need for second-order gradient calculations of DNNs.

\subsection{Coreset Method}

The coreset method reduces the size of the training dataset by selecting a subset of the original dataset that only includes representative samples. This approach can be categorized into geometric methods~\cite{sener2017active,chen2012super}, loss-based methods~\cite{toneva2018empirical}, and decision boundary-based methods~\cite{margatina2021active}. The geometric methods assume that data points close to each other in the feature space tend to have similar properties. Unlike geometric methods, the loss-based methods assume that training samples contributing more to the error or loss in neural network training are more important. The decision boundary-based methods assume that data points near the decision boundary are difficult to separate and can therefore be used as a coreset. However, all of these coreset methods rely on the forward inference results of DNNs to select representative samples. This reliance leads to reduced generalization ability due to the inductive bias errors of the DNNs. In contrast, our NMS employs DE-SNE to directly perform dimensionality reduction and sampling on the original images without the involvement of intermediate DNNs, thereby improving the generalization ability of NMS sampling.

\subsection{Limitation and Analysis of Previous Works}
\label{limit_analy}

1) \textbf{Poor Generalization:} Since the performance of synthetic or extracted representative images is evaluated using specific neural networks, the representative images obtained by DD and Coreset inevitably carry a preference for specific neural networks, leading to poor generalization. For example, the SOTA works of DD and Coreset, such as DM \cite{dm} and DQ \cite{dq}, show a significant drop in training accuracy when using images synthesized by ResNet18\cite{resnet} on ViT\cite{vit}, decreasing by 52.5\% (74.1\% vs. 21.6\%) and 30.1\% (82.7\% vs. 52.6\%), respectively.

2) \textbf{High Computational Cost for Representative Images:} The DD and Coreset methods generate representative images using DNNs, incurring substantial computational costs. For instance, DM requires 2800 GPU hours to generate 60\% of representative images. The SOTA works, \eg, DQ and DQAS, take 2.3 times longer to extract representative images compared to DNN training. Moreover, NeSSA has implemented a DNN forward inference accelerator on an FPGA, looping 200 times to input all image data for sampling. On edge devices, limited battery energy cannot support such substantial energy demands for generating representative images, making these methods impractical for edge device applications.

3) \textbf{DDR Energy Consumption Issue:} Besides computational energy costs, DDR data transfer energy is also a significant energy factor. For example, training a ResNet50 neural network on a GPU can have DDR energy consumption accounting for up to 40.4\% \cite{nessa}. However, previous architectures have neglected DDR energy optimization. For example, the NeSSA system~\cite{nessa}, most similar to our approach, requires multiple rounds of inputting all images into the accelerator.

\section{NMS}
\subsection{Challenge of Sampling with Data Generalization}

\begin{figure}[ht]
\centering
\includegraphics[width=0.7\columnwidth]{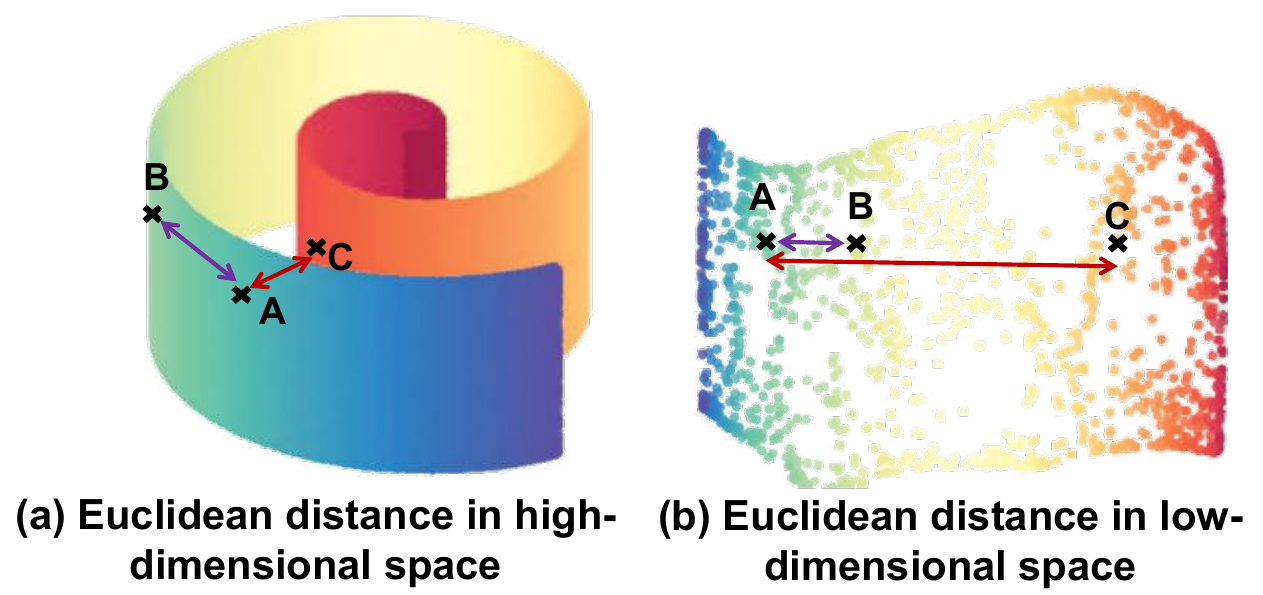}
\caption{The mismatch phenomenon is evident between samples using Euclidean distance measurement in different spaces. In high-dimensional space, the distance between samples A and B (AB) is greater than that between samples A and C (AC). However, in a low-dimensional manifold space, the distance AB is less than AC. The color represents the actual distance.}
\label{fig_dis_metric}
\end{figure}

As mentioned in the previous section, the DD and Coreset methods suffer from poor generalization due to sampling data with specific neural networks. To overcome this issue, it is necessary to choose a new sample sampling algorithm that selects training samples solely based on the data distribution of the images themselves. However, sampling in high-dimensional space introduces two significant challenges: (1) It is difficult to evaluate the similarity between samples in high-dimensional space. If we consider the data distribution of samples as shown in Fig.~\ref{fig_dis_metric}, using Euclidean distance to measure similarity can lead to large errors because the samples lie in a high-dimensional folded manifold space. (2) The computation of distances between samples in high-dimensional space is computationally intensive. For example, evaluating the distances between all images in the CIFAR10 dataset (resolution 32x32, 50,000 images) using Euclidean distance has a complexity of \((32 \times 32 \times 3) \times \binom{50000}{2} = O(10^{12})\), requiring trillions of computations.

To address these issues, we draw inspiration from the human brain mechanism for image processing and employ dimensionality reduction algorithms for image preprocessing. Manifold learning includes both linear and nonlinear manifold dimensionality reduction algorithms. Linear manifold dimensionality reduction algorithms include Principal Component Analysis (PCA) \cite{mackiewicz1993principal}, Linear Discriminant Analysis (LDA) \cite{xanthopoulos2013linear}, and Multi-Dimensional Scaling (MDS) \cite{petrick2002development}. Nonlinear manifold dimensionality reduction algorithms, inspired by the low-dimensional manifold stationary of the human brain, include Isometric Mapping (ISOMAP) \cite{samko2006selection}, Locally Linear Embedding (LLE) \cite{roweis2000nonlinear}, Laplacian Eigenmap (LE) \cite{belkin2003laplacian}, and t-distributed Stochastic Neighbor Embedding (t-SNE) \cite{tsne}. The t-SNE algorithm is a global dimensionality reduction algorithm that improves the asymmetry and crowding problems of the Stochastic Neighbor Embedding (SNE) algorithm, making it the optimal choice for this work.

\subsection{Challenge of Sampling with t-SNE Algorithm}
\begin{figure}[ht]
    \centering
    \includegraphics[width=0.5\columnwidth]{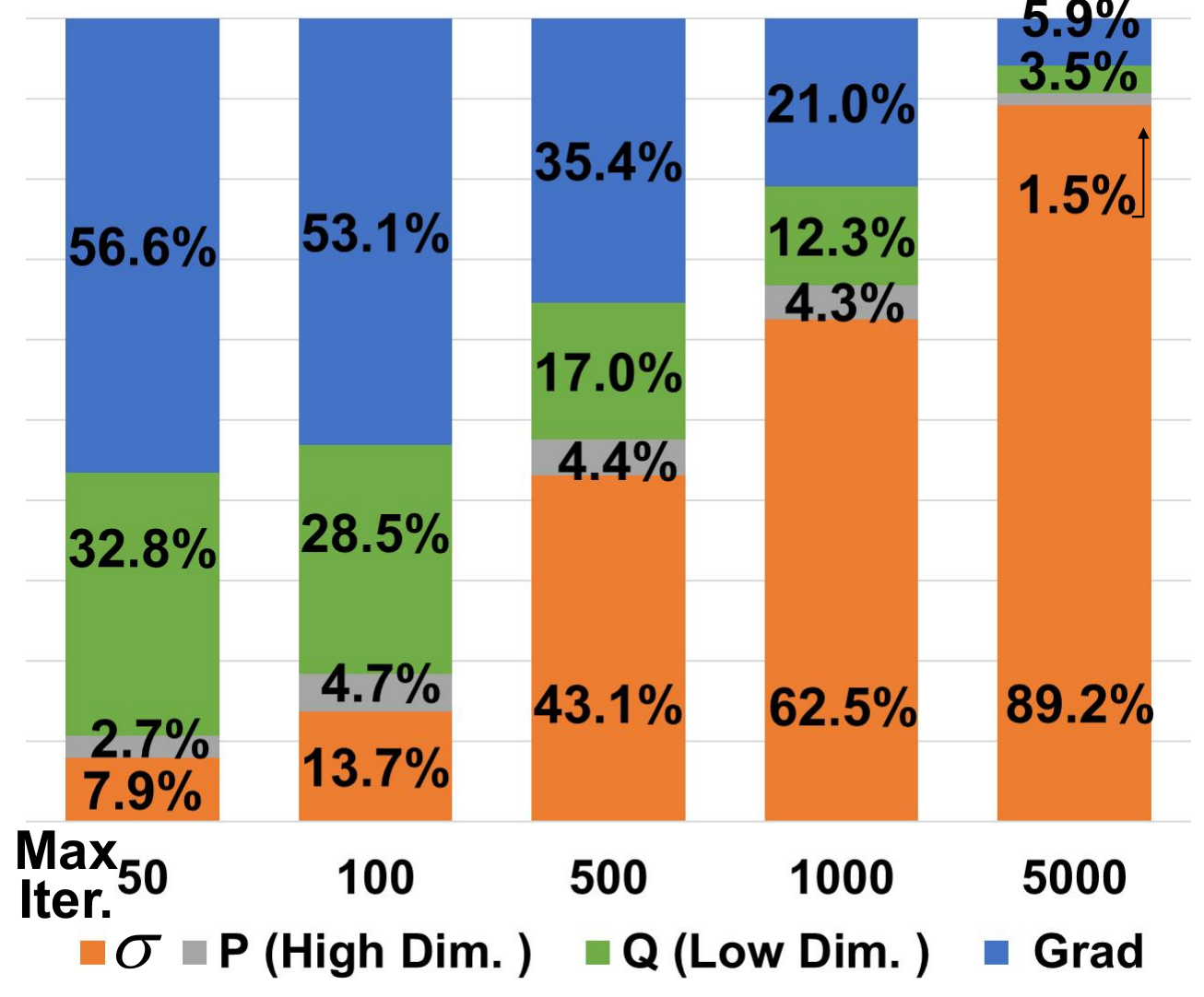} 
    \caption{Breakdown of the running times for various operations in the t-SNE. '$\sigma$' adjusts to find the optimal perplexity. 'P' and 'Q' represent high- and low-dimensional probability distributions, respectively. 'Grad' solves the low-dimensional embedding via gradient descent.
    }
    \label{fig_tsne_sigma_breakdown} 
\end{figure}

Although previous studies have demonstrated that the t-SNE algorithm offers optimal performance, it requires significant computational resources \cite{tsne}. To facilitate better deployment on edge devices, we analyze the runtime of each component of the algorithm to identify bottlenecks. As shown in Fig.~\ref{fig_tsne_sigma_breakdown}, as the scale of the search problem increases, such as the increase in image size and the number of images, the number of iterations in the bisection method increases. Consequently, the time spent searching for the parameter \(\sigma\) becomes increasingly significant. For instance, when the number of iterations reaches 5000, the time spent searching for \(\sigma\) can account for up to 90\% of the total time.

\label{chall_t-SNE}

\begin{figure}[h]
    \centering
    \includegraphics[width=0.7\columnwidth]{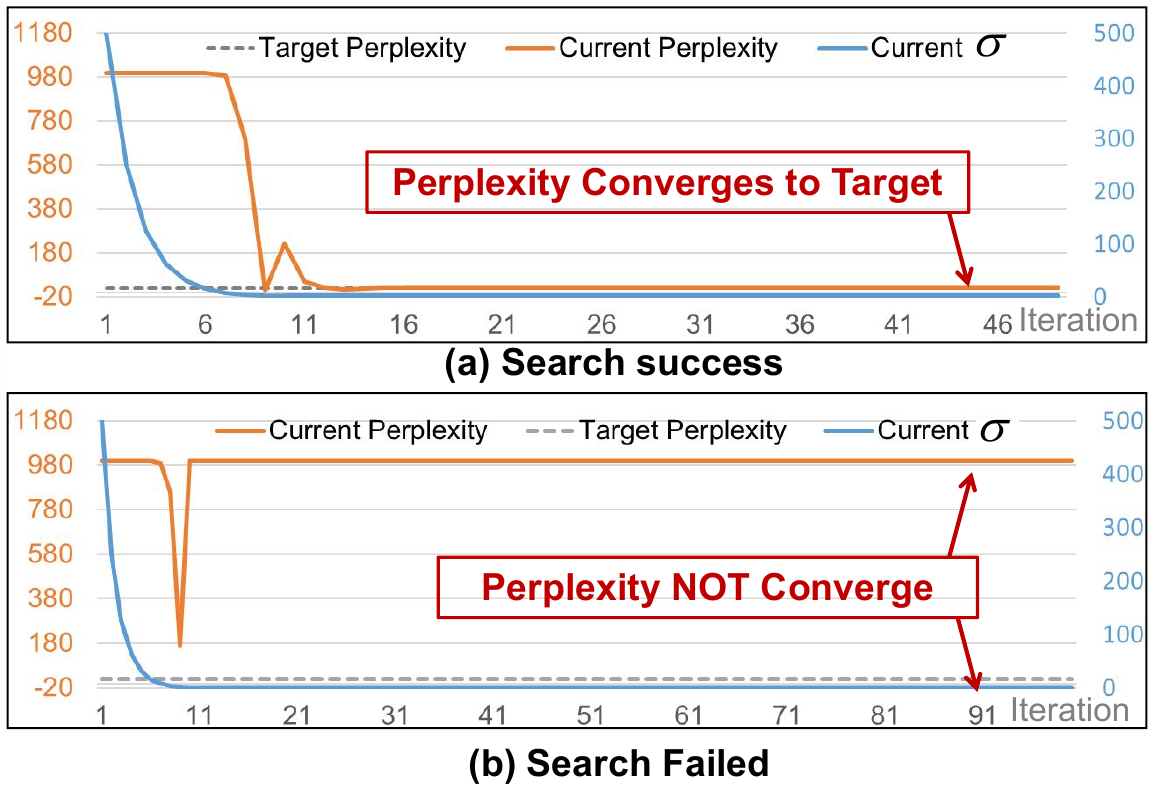} 
    \caption{Searching for the optimal \(\sigma\) to achieve the desired perplexity in the conventional t-SNE algorithm.} 
    \label{fig_sigma_failed}
\end{figure} 

To address this challenge, we conduct a detailed analysis of the reasons for the prolonged search for the \(\sigma\) parameter. The primary purpose of the \(\sigma\) parameter search is to find the optimal value of its function value perplexity. For the first time, we discover that the function value perplexity sometimes surpasses the optimal value with changes in the \(\sigma\) parameter, causing the algorithm to enter ineffective iterations and fail to converge, as shown in Fig.~\ref{fig_sigma_failed}. This results in bigger errors in perplexity values. In other words, the perplexity search in t-SNE is a continuous non-monotonic optimization problem, whereas the binary search method is only suitable for discrete monotonic problems. Therefore, the binary search method may not be suitable for this issue. To address this, we propose an perplexity search algorithm based on Differential Evolution (DE) \cite{de} and combine the t-SNE sampling algorithm, termed DE-SNE.
\begin{figure}[ht]
\centering
\includegraphics[width=1.0\columnwidth]{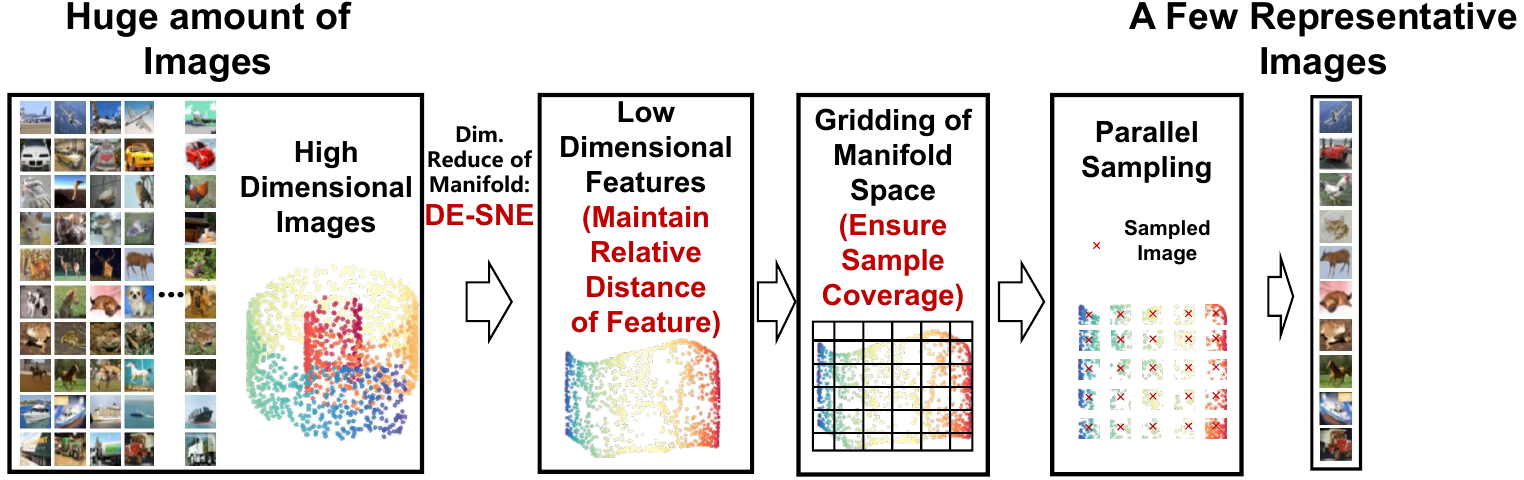} 
\caption{Illustration of the brain-inspired DE-SNE dataset sampling compression method.}
\label{fig_sample_flow}
\end{figure} 

\begin{figure*}[ht]
\centering
\includegraphics[width=1.3\columnwidth]{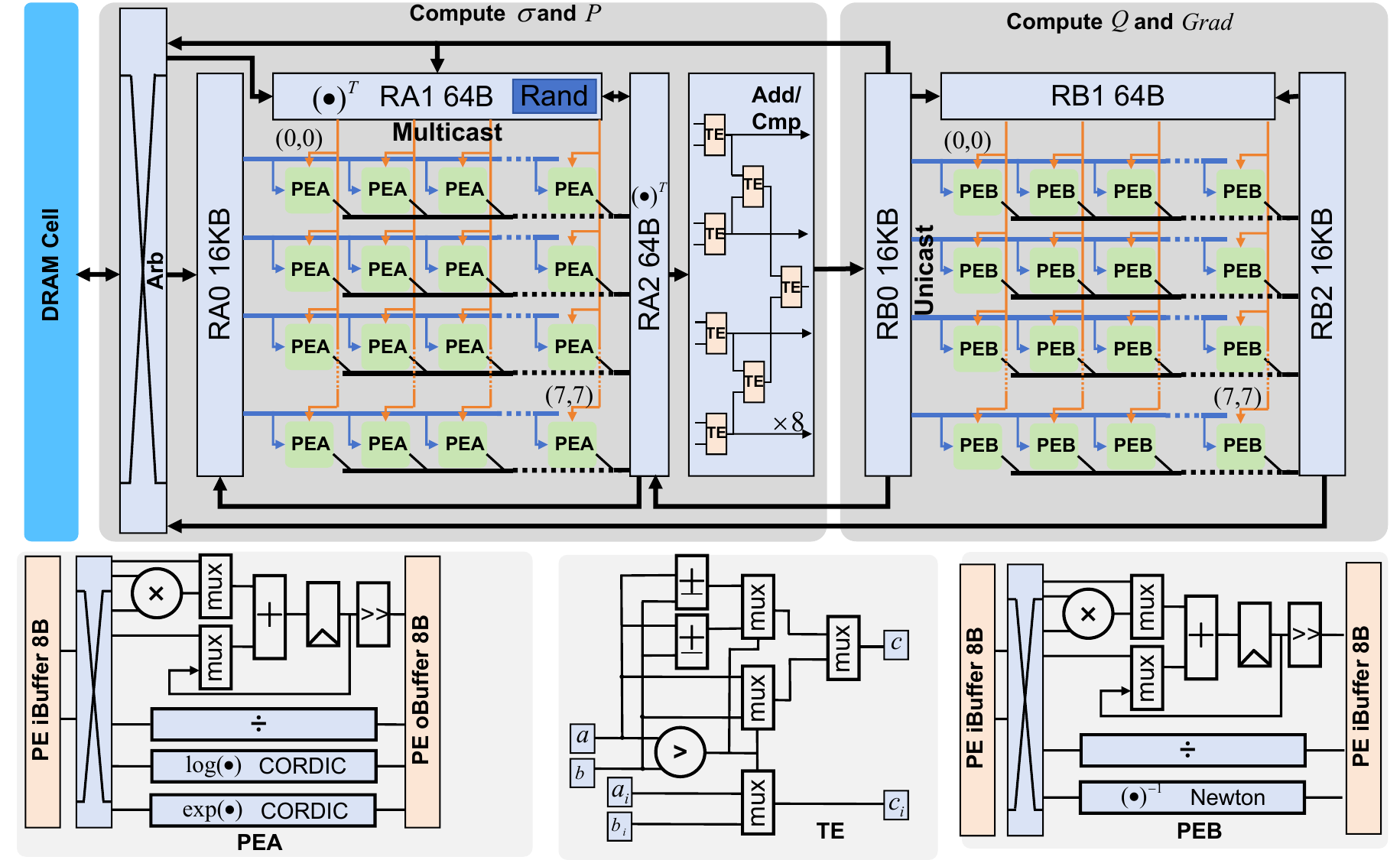} 
\caption{The accelerator of DE-SNE.}
\label{fig_tsne_hw}
\end{figure*} 
There have been some designs for specialized accelerators for the DE algorithm, such as \cite{de_hw0,de_hw1,de_hw2,de_hw3,de_hw4}. However, using the DE algorithm for perplexity function value search in edge-side accelerators is extremely challenging. Previous works on DE accelerators can only accelerate the DE algorithm itself and can't perform other steps in the t-SNE algorithm, such as calculating the P matrix, Q matrix, and gradient descent. Although we can design another specialized accelerator for other operators, this results in greater circuit energy and area overhead, which is not acceptable for edge devices with strict efficiency requirements. Thus, designing an efficient accelerator that can accommodate all steps of the t-SNE algorithm (including DE operators) in an edge environment remains a significant challenge.

\subsection{DE-SNE Sampling Algorith}
We first propose an image sampling scheme based on DE-SNE, with specific steps as follows:
\begin{enumerate}
    \item Nonlinear manifold dimensionality reduction;
    \item Gridding of low-dimensional manifold space;
    \item Parallel sampling.
\end{enumerate}

Specifically, the nonlinear manifold dimensionality reduction is primarily implemented using the DE-SNE algorithm, which preserves the relative distances of samples in the high-dimensional space, ensuring the representativeness of the subsequent sampled samples. The gridding of the low-dimensional manifold space mainly ensures good coverage of the sampled samples across the entire feature space. In parallel sampling, a certain number of representative samples are randomly selected within each grid to form the final condensed dataset. For the detail illustrate please refer to Fig.~\ref{fig_sample_flow}. Although the DE-SNE algorithm has good manifold stationary, it has a large computational workload compared to other dimensionality reduction algorithms, necessitating the development of efficient specialized sampling circuits.

\subsection{DE-SNE Sampling Circuit}

 As shown in Fig.~\ref{fig_tsne_hw}, the DE-SNE sampling circuit consists mainly of three parts: the Processing Element A Array (PEA Array), the Tree Element Array (TE Array), and the Processing Element B Array (PEB Array). The PEA Array, composed of 64 PEAs, primarily calculates the \(\sigma\) and P matrices. The TE Array, consisting of 8 groups of 64 TEs, mainly handles vector comparison and addition. The PEB Array including 64 PEBs calculates the Q matrix and gradients. The PEA module employs the CORDIC method to implement logarithmic and exponential operations, Newton's method \cite{newton_neg_one_pow} for power of -1 operations, and also performs multiplication and addition for vector and matrix operations. The PEB uses a similar circuit structure to the PEA, as it does not need to implement nonlinear operations. In addition, the DE-SNE sampling circuit includes three 64 KB SRAMs: RA0, RB0, and RB2, and three 64 B SRAMs: RA1, RA2, and RB1.

\begin{figure}[ht]
\centering
\includegraphics[width=1.0\columnwidth]{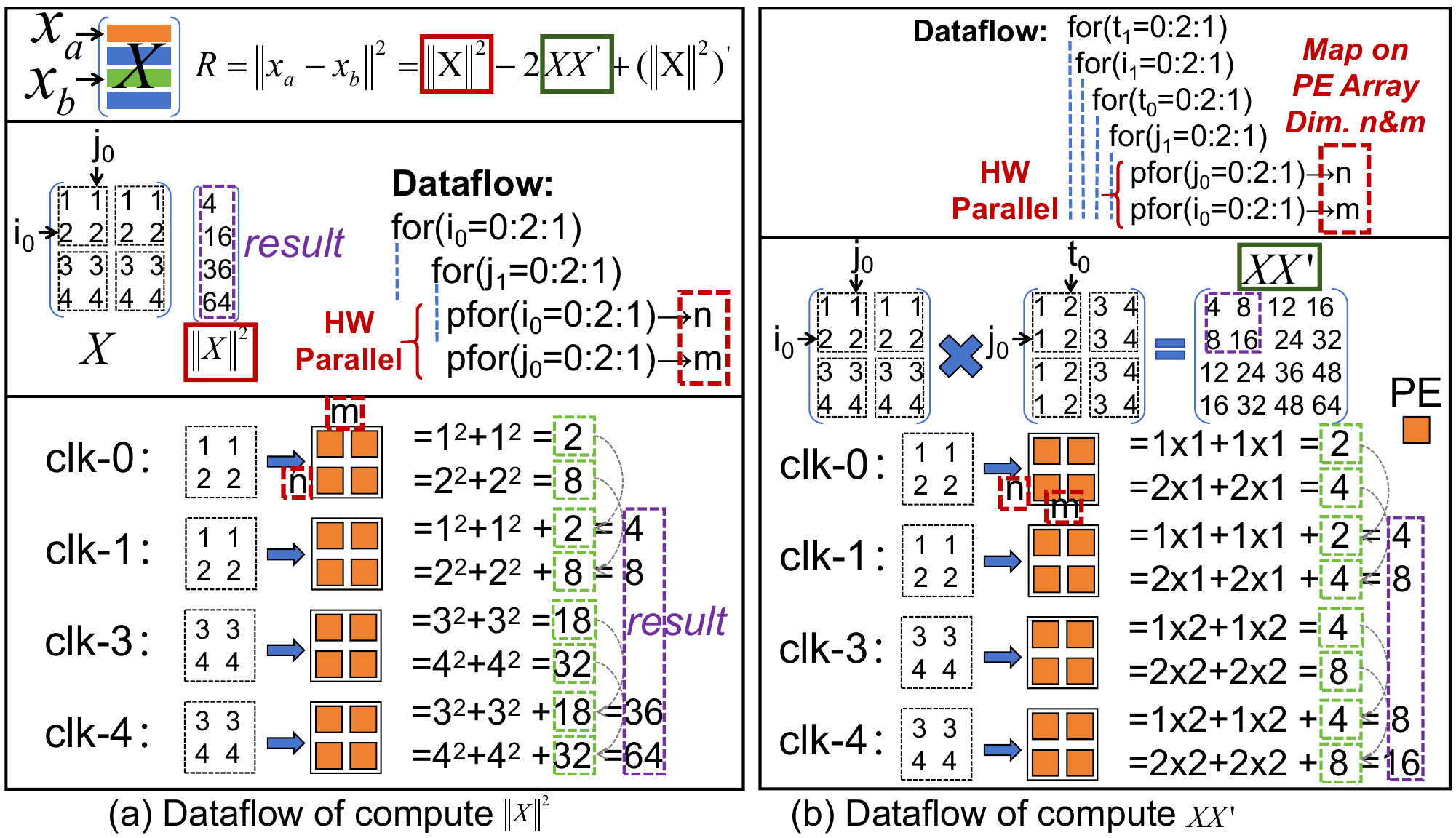} 
\caption{Dataflow of calculating distance matrix.}
\label{fig_dataflow_dis}
\end{figure} 

\section{Key Operations of NMS Data Flow}
\subsection{Distance Matrix Calculation}
In the workflow of the DE-SNE algorithm, most operators can be decomposed into common matrix and vector operations. These operations can then be mapped onto the computational architecture shown in Fig.~\ref{fig_dataflow_dis}, enabling parallel implementation of the DE-SNE algorithm using a uniform hardware architecture. However, there is a class of operators that cannot be directly accelerated using this regular architecture. Specifically, the calculation of the distance matrix $\|\mathbf{x}_a - \mathbf{x}_b\|^2$, as shown in Eq. (\ref{math_pji}) and (\ref{math_qij}), requires computing the L2 norm distance between arbitrary pairs of feature vectors. The input is an $N \times D_i$ matrix, and the output is a $D_i \times D_i$ square matrix, which does not correspond to a typical linear algebra operation. Thus, we adopt the transformation shown in Fig.~\ref{fig_dataflow_dis}(a), where the distance matrix is decomposed into the sum and difference of three matrices, namely $\|\mathbf{x}\|^2$ and $\mathbf{x}\mathbf{x}^T$. The lower-middle part of Fig.~\ref{fig_dataflow_dis}~(a) illustrates how to compute $\|\mathbf{x}\|^2$ using traditional matrix multiplication. Fig.~\ref{fig_dataflow_dis}~(b) demonstrates the computation of $\mathbf{x}\mathbf{x}^T$. For efficient computation, we propose a data flow architecture compatible with a uniform structure, which efficiently utilizes the PEA array to compute the distance matrix.

\subsection{DE Algorithm Calculation}
As detailed in Section~\ref{chall_t-SNE}, implementing the DE algorithm on a uniform hardware architecture is challenging due to branching and jump instructions, such as in lines 13, 16, and 18 of Algorithm~\ref{alg_de}. 

\begin{algorithm} \small
\caption{Differential Evolution (DE)}
\begin{algorithmic}[1]
\REQUIRE $f$, $tgt$, $\epsilon = 1 \times 10^{-10}$, $max\_iter = 10000$, $pop\_size = 30$, $lb = 1 \times 10^{-20}$, $ub = 1000$, $F = 0.5$, $CR = 0.7$
\ENSURE $best\_ind$
\STATE Initialize $pop \gets \text{Uniform}(lb, ub, pop\_size)$
\STATE $best\_ind \gets pop[\arg\min(|f(ind) - tgt| \text{ for } ind \text{ in } pop)]$
\STATE $best\_val \gets f(best\_ind)$

\FOR{$iter = 1$ \TO $max\_iter$}
    \FOR{$i = 1$ \TO $pop\_size$}
        \STATE \textbf{Mutation:}
        \STATE $idxs \gets [j \text{ for } j \in [0, pop\_size) \text{ if } j \neq i]$
        \STATE $a, b, c \gets \text{Random choice from } idxs$
        \STATE $mut \gets a + F \cdot (b - c)$
        \STATE $mut \gets \text{clip}(mut, lb, ub)$
        
        \STATE \textbf{Crossover:}
        \STATE $c\_rand \gets \text{Random number in } [0, 1]$
        \STATE $trial \gets \text{if } c\_rand < CR \text{ then } mut \text{ else } pop[i]$
        
        \STATE \textbf{Selection:}
        \STATE $trial\_val \gets f(trial)$
        \IF{$|trial\_val - tgt| < |f(pop[i]) - tgt|$}
            \STATE $pop[i] \gets trial$
            
            \IF{$|trial\_val - tgt| < |best\_val - tgt|$}
                \STATE $best\_ind \gets trial$
                \STATE $best\_val \gets trial\_val$
            \ENDIF
        \ENDIF
    \ENDFOR
\ENDFOR

\RETURN $best\_ind$
\end{algorithmic}
\label{alg_de}
\end{algorithm}

\begin{figure}[ht]
\centering
\includegraphics[width=0.70\columnwidth]{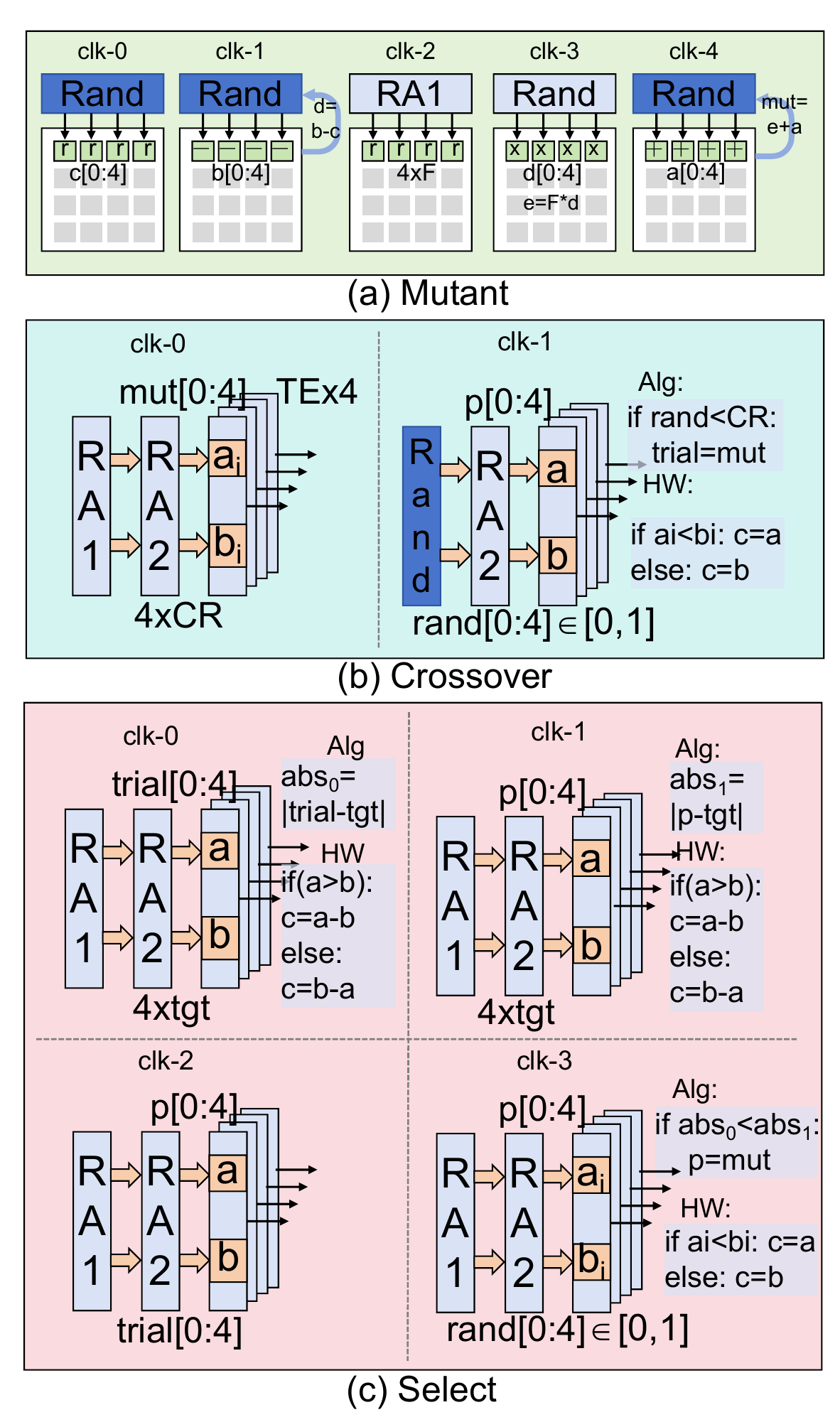} 
\caption{Dataflow of processing differential evolution.}
\label{fig_de_dataflow}
\end{figure} 

To overcome this, we first set up 32 individuals for parallel execution at the algorithm level. At the hardware level, we reuse the comparators in the TE module, allowing the accelerator to perform parallel computations across individuals. Specifically, different hardware components are employed for mutation, crossover, and selection operations. For the mutation operation, three individuals, $a$, $b$, and $c$, are randomly generated, followed by the generation of a mutant individual in line 9. As illustrated in Fig.\ref{fig_de_dataflow}~(a), four sets of random numbers are generated using the pseudo-random number generation module in RA1 to select $a[0:4]$, $b[0:4]$, and $c[0:4]$. The PE unit in the PEA matrix then calculates the mutation result, $mut$.

For the crossover operation, as shown in line 8 and Fig.\ref{fig_de_dataflow}~(b), a random number between $[0,1]$ is generated as the crossover probability. Based on this probability, either the mutant individual $mut$ or the original individual $p[i]$ is selected as $trial[i]$. The TE module then performs the selection operation in parallel. As depicted in Fig.\ref{fig_de_dataflow}~(c), the fitness function values determine whether to select the crossover individual $trial$ as the new individual, updating the record of the best individual. This process also utilizes the TE module for parallel comparison.

\section{Near-Memory Computing Architecture}

\begin{figure}[ht]
\centering
\includegraphics[width=0.9\columnwidth]{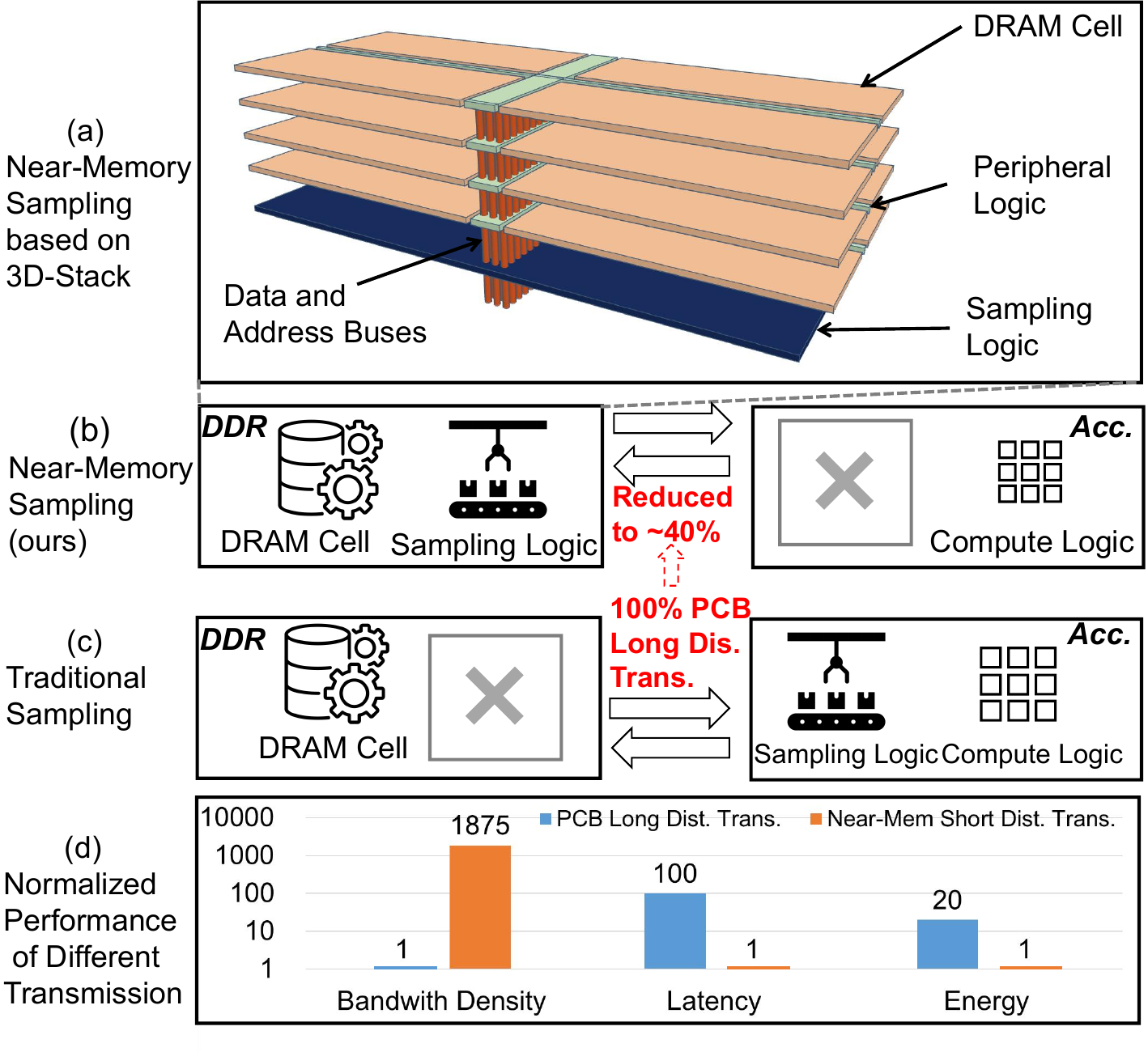} 
\caption{Near-memory sampling architecture.}
\label{fig_nms_arch}
\end{figure}

Section~\ref{limit_analy} discusses that the energy overhead from data transmission between DDR and the accelerator can exceed 40\% during DNNs training, significantly impacting the deployment of edge devices. While traditional data sampling algorithms theoretically reduce the number of training images, in practical deployment, all training images still need to pass through the image sampling circuit to identify which images should be retained for actual model training. This process significantly increases DDR transmission and the corresponding energy overhead, as shown in Fig.~\ref{fig_nms_arch}(d) and Fig.~\ref{fig_nms_arch}(c).

To address this issue, we introduce a Near-Memory Sampling architecture, integrating image-sampling logic circuits with DDR near-memory computing. As shown in Fig.~\ref{fig_nms_arch}(a) and Fig.~\ref{fig_nms_arch}(b), the sampling logic circuit and DRAM cells are packaged together using 3D stacking. Previous research~\cite{sharma2022universal,ajanovic2008pci} suggests that this architecture can reduce the energy consumption per transmitted bit by 20 times, reduce transmission latency by 100 times, and increase data transmission bandwidth density by over three orders of magnitude. This architecture greatly reduces the data transmission load over long-distance PCBs, transmitting only representative image samples filtered by the DE-SNE algorithm to the training accelerator, thereby significantly reducing the energy overhead for edge device training.

Notably, implementing the sampling logic circuit within a near-memory computing architecture appears straightforward but hinges on the ability to employ the DE-SNE algorithm for data sampling without relying on DNN inference. Previous works~\cite{nessa,dq}, reliant on DNN inference for image extraction, have found it challenging to adopt similar schemes. 

%

\section{Experiments and Analysis}

\subsection{Experimental Setup}

\textbf{Datasets:} Following the experimental setups outlined in the literature \cite{nessa,dq,Zhao2024b}, we selected a total of six datasets: CIFAR10, CINIC10, SVHN, CIFAR100, ImageNet-100, and ImageNet-1K. To evaluate the performance of the sampling training algorithm across datasets of varying scales, we categorized the datasets into small-scale and large-scale groups. The small-scale datasets, comprising CIFAR10, CINIC10, and SVHN, each contain 10 classes with image dimensions of $32 \times 32 \times 3$. The large-scale datasets include CIFAR100, ImageNet-100, and ImageNet-1K, where CIFAR100 and ImageNet-100 consist of 100 classes with image sizes of $32 \times 32 \times 3$ and $224 \times 224 \times 3$, respectively, while ImageNet-1K encompasses 1000 classes with an image dimension of $224 \times 224 \times 3$. In our comparative analysis against state-of-the-art (SOTA) works DQ \cite{dq}, DQAS \cite{Zhao2024b}, and NeSSA \cite{nessa}, we note that the DQ and DQAS methods only provide implementations for CIFAR10. Therefore, we have reproduced the DQ and DQAS experiments across other datasets for a comprehensive comparison.


\textbf{Models:} Following previous works~\cite{dq,Zhao2024b,dm,nessa}, we utilize the PyTorch framework to evaluate a range of DNN and Transformer networks, including ResNet18~\cite{resnet}, ResNet50~\cite{resnet}, ShuffleNetV2~\cite{shufflenetv2}, MobileNetV2~\cite{sandler2018mobilenetv2}, and ViT~\cite{vit}. We use an SGD optimizer with a momentum of 0.9 in 200 epochs, applying a batch size of 64 for small-scale datasets and 128 for large-scale datasets. A cosine-annealed learning rate schedule is employed, with an initial learning rate of 0.1. The sampling keeping ratios are set at 10\%, 20\%, and 30\% to evaluate the model performance.

\textbf{Hardware:} The server utilized for the experiments is equipped with an AMD $\texttt{EPYC}^{TM}$ 7H12 64-Core Processor and six NVIDIA GeForce RTX 4090 GPUs, with memory of 512GB. To evaluate our data flow and hardware architecture, we develop a cycle-accurate simulator to calculate latency and memory traces. The hardware architecture is implemented in Verilog RTL and synthesized using Synopsis Design Compiler with TSMC 28 nm technology. Additionally, on-chip buffer energy consumption is estimated using CACTI 7 \cite{balasubramonian2017cacti}. The energy consumption for near-memory access is set to 0.5 pj/bit, using data from the open UCIe protocol \cite{sharma2022universal}. For board-level DDR access, the energy consumption is set to 10 pj/bit, based on PCB-level power data from the PCIe protocol \cite{ajanovic2008pci}. To ensure a fair comparison, we standardize the per-byte PCB-level DDR access energy consumption across NeSSA, DQ, and our method. The DE-SNE sampling circuit is integrated with the current SOTA sparse floating-point training accelerator THATA~\cite{theta} into a unified system and compares with a system that doesn't use the sample sampling circuitry. Due to the lack of the original algorithm implementation in THATA, this experiment only compared hardware efficiency.

\begin{table}[]
    \setlength\tabcolsep{2.0pt}   
    \centering  
    \caption{The Top-1 accuracy comparison with SOTA Coreset methods.}		 
    \begin{tabular}{@{}l|c|c|ccccc|c@{}}
    \toprule
Dataset & Method                                 & KR(\%) & R18                 & R50           & ShfV2         & MbV2          & ViT           & Avg.(\%)     \\ \cline{2-9}
\multirow{13}{*}{CIFAR10} & Full data             & 100    & 95.6                & 95.5          & 85.0          & 90.2          & 80.2         & 89.3         \\ \cline{2-9} 
                          & \multirow{3}{*}{DQ}  & 10     & 85.2                & 81.4          & 73.4          & 74.5          & 52.6          & 73.4         \\
                          &                      & 20     & 89.4                & 84.9          & 81.6          & 81.4          & 65.9          & 80.6         \\
                          &                      & 30     & 91.8                & 89.9          & 81.8          & 86.0          & 71.3          & 84.2         \\ \cline{2-9} 
                          & \multirow{3}{*}{DQAS}& 10     & 86.1                & 80.8          & 72.3          & 75.7          & 53.5          & 73.7         \\
                          &                      & 20     & 90.2                & 85.1          & 80.5          & 80.1          & 66.8          & 80.5         \\
                          &                      & 30     & 93.3                & 87.8          & 81.2          & 87.2          & 72.2          & 84.3         \\ \cline{2-9} 
                          & \multirow{3}{*}{NeSSA}& 10    & 87.8                & 81.3          & 72.5          & 76.8          & 58.1          & 75.3         \\
                          &                      & 20     & 88.7                & 85.5          & 80.7          & 84.4          & 69.8          & 81.8         \\
                          &                      & 30     & 90.5                & 88.2          & 82.8          & 87.4          & 73.6          & 84.5         \\ \cline{2-9} 
                    & \multirow{3}{*}{NMS(ours)} & 10  & \textbf{88.0} & \textbf{82.4} & \textbf{75.7} & \textbf{80.4} & \textbf{61.9} & \textbf{ 77.7}         \\
                    &                            & 20  & \textbf{90.4} & \textbf{85.7} & \textbf{82.4} & \textbf{86.4} & \textbf{70.3} & \textbf{ 83.0}         \\
                    &                            & 30  & \textbf{93.5} & \textbf{90.0} & \textbf{84.0} & \textbf{88.3} & \textbf{75.6} & \textbf{ 86.3}         \\ \hline
\multirow{13}{*}{CINIC10} & Full data            & 100    & 82.5                & 81.7          & 74.7          & 78.7          & 76.8          & 78.9          \\ \cline{2-9}
                          & \multirow{3}{*}{DQ}  & 10     & 68.9                & 64.1          & 61.5          & 62.8          & 37.4          & 58.9          \\
                          &                      & 20     & 75.9                & 76.6          & 67.8          & 70.9          & 49.0          & 68.0          \\
                          &                      & 30     & 79.8                & 80.3          & 70.0          & 75.1          & 55.3          & 72.1          \\  \cline{2-9}
                          & \multirow{3}{*}{DQAS}& 10     & 68.0                & 63.1          & 60.2          & 64.1          & 37.6          & 58.6          \\
                          &                      & 20     & 72.9                & 76.2          & 66.7          & 69.8          & 50.5          & 67.2          \\
                          &                      & 30     & 77.0                & 77.9          & 69.1          & 76.4          & 56.8          & 71.4          \\  \cline{2-9}
                          & \multirow{3}{*}{NeSSA}& 10    & 69.2                & 63.8          & 61.4          & 65.8          & 40.1          & 60.1          \\
                          &                      & 20     & 75.7                & 73.6          & 67.8          & 70.5          & 51.3          & 67.8          \\
                          &                      & 30     & 80.3                & 78.8          & 70.8          & 76.8          & 63.6          & 74.1          \\ \cline{2-9}
                    & \multirow{3}{*}{NMS(ours)} & 10  & \textbf{70.8}  & \textbf{65.1} & \textbf{63.5} & \textbf{69.2} & \textbf{46.4} & \textbf{63.0}         \\
                          &                      & 20  & \textbf{77.2}  & \textbf{77.0} & \textbf{68.5} & \textbf{76.3} & \textbf{53.3} & \textbf{70.5}         \\
                          &                      & 30  & \textbf{81.1}  & \textbf{80.4} & \textbf{72.4} & \textbf{77.5} & \textbf{69.9} & \textbf{76.3}         \\  \hline
\multirow{13}{*}{SVHN}    & Full data             & 100    & 95.9               & 96.1          & 95.2          & 95.7          & 52.6          & 87.1          \\ \cline{2-9}
                          & \multirow{3}{*}{DQ}  & 10     & 92.5                & 89.1          & 90.3          & 87.5          & 10.2          & 73.9          \\
                          &                      & 20     & 93.6                & 92.1          & 93.0          & 89.2          & 26.4          & 78.9          \\
                          &                      & 30     & 94.8                & 94.8          & 93.4          & 93.6          & 27.4          & 80.8          \\ \cline{2-9}
                          & \multirow{3}{*}{DQAS}& 10     & 92.0                & 89.4          & 89.6          & 89.4          & 11.0          & 74.3          \\
                          &                      & 20     & 90.8                & 91.4          & 92.5          & 88.4          & 27.9          & 78.2          \\
                          &                      & 30     & 92.6                & 92.4          & 91.5          & 94.7          & 29.1          & 80.1          \\ \cline{2-9}
                          & \multirow{3}{*}{NeSSA}& 10    & 92.9                & 90.1          & 90.5          & 89.6          & 19.5          & 76.5          \\
                          &                      & 20     & 95.2                & 91.9          & 92.8          & 90.7          & 29.8          & 80.1          \\
                          &                      & 30     & 95.3                & 93.4          & 92.9          & 94.6          & 40.8          & 83.4          \\ \cline{2-9}
                    & \multirow{3}{*}{NMS(ours)} & 10  & \textbf{93.8}  & \textbf{90.4} & \textbf{92.7} & \textbf{93.6} & \textbf{19.6} & \textbf{78.0}         \\
                          &                      & 20  & \textbf{95.4}  & \textbf{92.3} & \textbf{94.1} & \textbf{94.6} & \textbf{30.8} & \textbf{81.4}         \\
                          &                      & 30  & \textbf{95.6}  & \textbf{95.1} & \textbf{94.9} & \textbf{95.2} & \textbf{41.2} & \textbf{84.4}         \\ 
\bottomrule
\end{tabular}
\label{exp_vs_coreset}
\end{table}


\textbf{Software:} To compare the efficiency and overhead of various perplexity optimization methods, we select common optimization algorithms, including binary search (BS) \cite{bs}, simulated annealing (SA) \cite{sa}, genetic algorithms (GA) \cite{ga}, particle swarm optimization (PSO) \cite{pso}, ant colony optimization (ACO) \cite{aco}, and DE algorithms (DE) \cite{de}. These algorithms are employed to solve the \(\sigma\) parameters on the above datasets, obtaining predicted perplexity values and comparing them with the input ground truth values to calculate the error. In this paper, we set the target perplexity to 15, a value determined through experimentation with different parameter configurations and informed by engineering experience.

\section{Evaluation of Algorithm Generalization}
\subsection{Comparison with SOTA Coreset Selection Methods}

As presented in TABLE~\ref{exp_vs_coreset}, compared to the current SOTA coreset sampling methods, e.g., DQ of ICCV-23, DQAS of ECCV-24, and NeSSA of HotStorage-23, our NMS achieves an average accuracy improvement of 2.5\% relative to DQ, 3.4\% relative to DQAS, and 2.8\% relative to NeSSA, with a maximum improvement of 9.3\% for DQ, 14.6\% for DQAS, and 13.8\% for NeSSA. These results highlight its effectiveness across different keeping ratios (KR) on the CIFAR10, CINIC10, and SVHN datasets, in both classical convolutional neural networks and vision transformer networks. We analyze that the improvement primarily comes from the use of our refined sampling algorithm without DNNs. Specifically, we exploit the DE-SNE algorithm that avoids the bias introduced by specific DNN models in the sampling process, enhancing the generalization ability of the sampled datasets.

\begin{figure*}[ht]
    \centering
    \includegraphics[width=1.95\columnwidth]{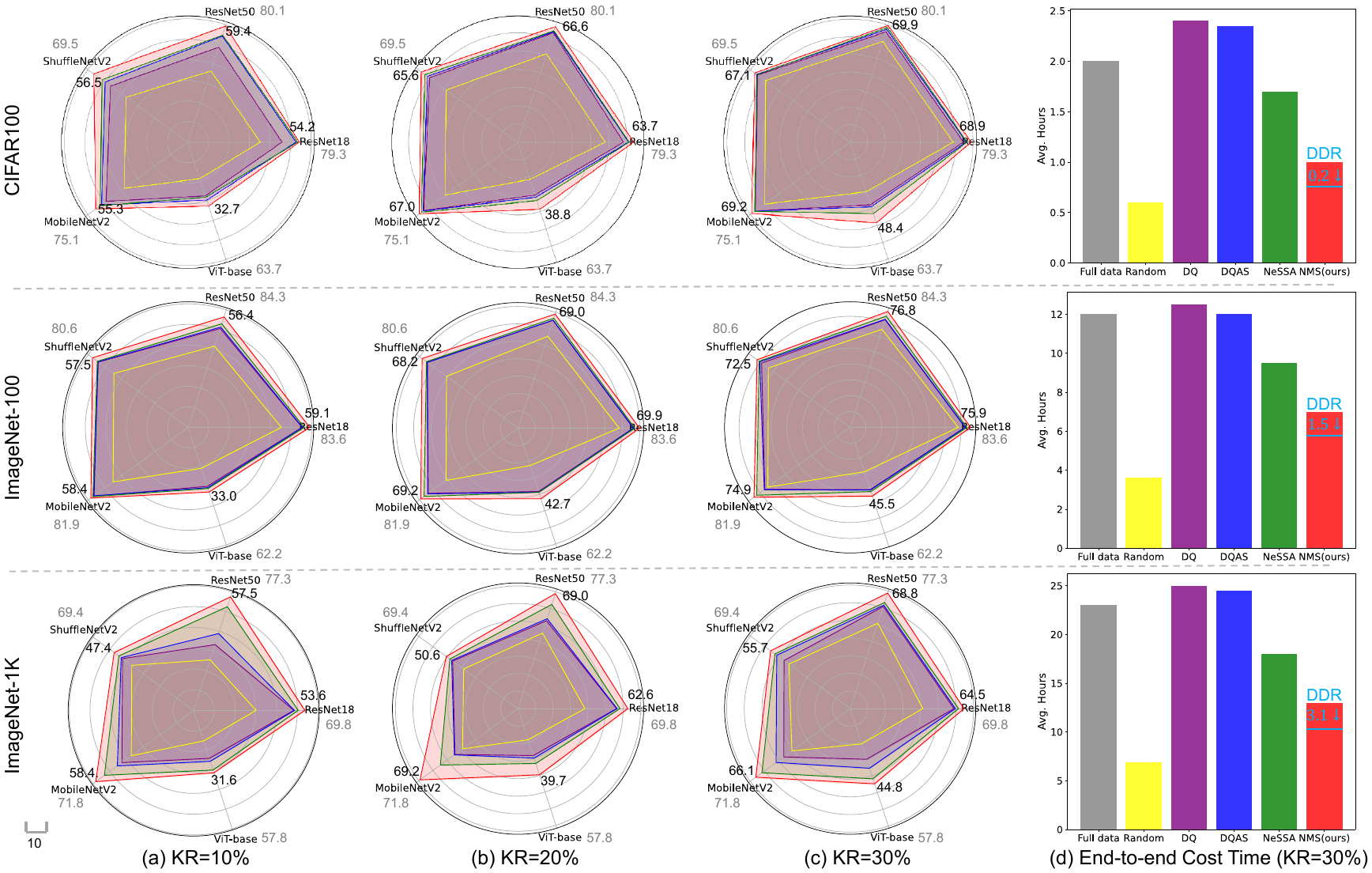}
    \caption{(a), (b) and (c): The Top-1 accuracy comparison with the SOTA coreset methods in different keeping ratios on complex datasets. (d): End-to-end time cost of ResNet18 on different datasets. The ``Full data" denotes the reproduced results using Pytorch's official \protect\href{https://github.com/pytorch/examples/tree/main}{code} in full dataset.}
    \label{exp_vs_coreset_radar}
\end{figure*} 
\begin{figure*}[ht]
\centering
\includegraphics[width=2.0\columnwidth]{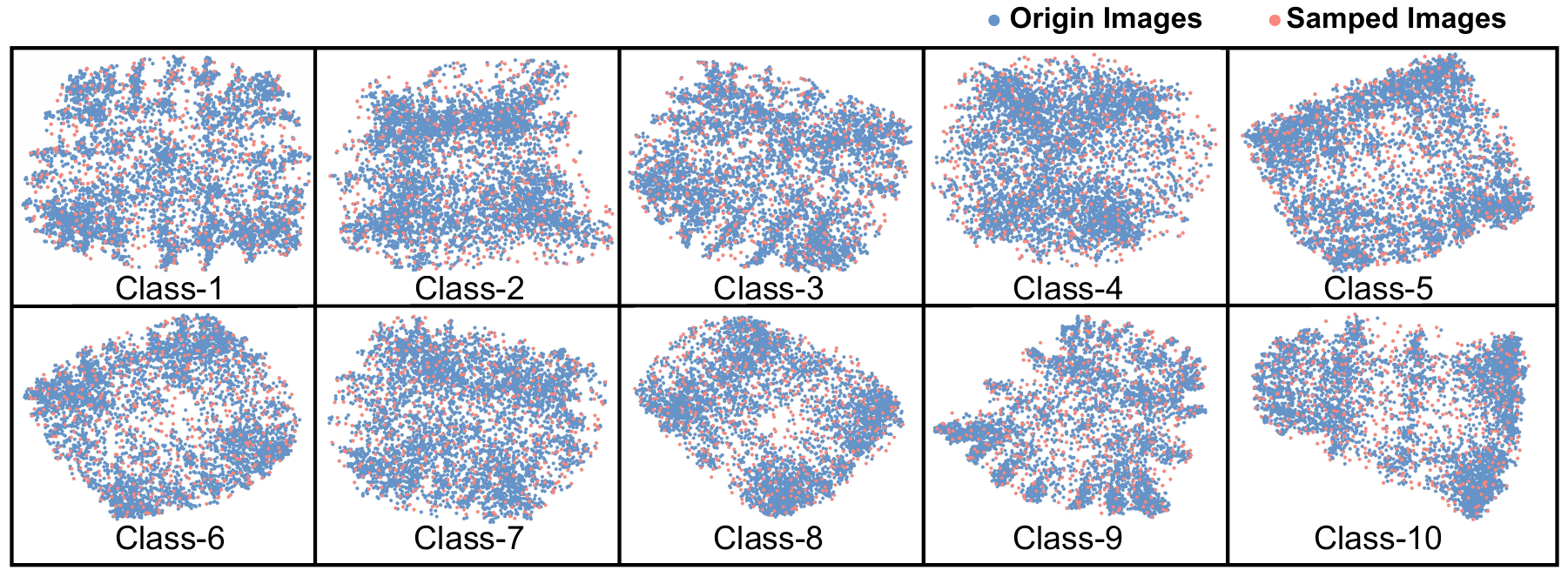} 
\caption{The visual results of our improved DE-SNE algorithm in the CIFAR10 dataset with a 10\% keeping ratio.} 
\label{fig_vis}
\end{figure*}

In addition, we conduct experiments on more complex datasets, namely CIFAR100, ImageNet-100, and ImageNet-1K, as shown in Fig.~\ref{exp_vs_coreset_radar}. The results consistently demonstrate that our NMS method outperforms other state-of-the-art (SOTA) coreset methods, proving its superiority and scalability. Across the CIFAR100 dataset, NMS achieves an average improvement of 6.0\% in Top-1 accuracy over DQ, 3.9\% over DQAS, and 3.0\% over NeSSA. The maximum observed improvements are 10.9\%, 9.7\%, and 5.5\%, respectively. On the ImageNet-100 dataset, NMS demonstrates average accuracy gains of 4.2\% over DQ, 3.9\% over DQAS, and 2.6\% over NeSSA. The maximum accuracy improvements are 8.4\%, 8.0\%, and 3.9\%, respectively. These results highlight the effectiveness of NMS, especially in comparison to DQ and DQAS, where it achieves more stable and consistent improvements across keeping ratios. NMS achieves an even greater impact for the more complex ImageNet-1K dataset, with average accuracy improvements of 11.9\% over DQ, 9.7\% over DQAS, and 4.7\% over NeSSA. The maximum improvements are 24.7\%, 24.3\%, and 14.4\%, respectively, demonstrating the scalability of NMS to large-scale datasets where performance improvements are crucial. When comparing NMS to the Random sampling baseline, the average accuracy improvement across all datasets and keeping ratios is 18.4\%, with a maximum observed improvement of 31.9\%. These substantial gains validate the superiority of NMS in providing unbiased sampling, which leads to better generalization and higher accuracy in both small and large datasets.

Moreover, regarding training efficiency, NMS consistently requires less training time than NeSSA and DQAS across all datasets and maintains the same ratios. Specifically, NMS is 6.3 times faster than DQ, 6.0 times faster than DQAS, and 2.7 times faster than NeSSA. Additionally, if we consider that NMS's sampling circuit and training circuit can operate in parallel on actual DDR hardware, NMS could achieve a further reduction of 24.6\%. In contrast, other algorithms, which rely on GPU for sampling, cannot parallelize the sampling and training processes. This highlights the parallelization advantages of the proposed DNN-free method presented in this study.


\subsection{Visualization of Sampled Data}
To intuitively understand the dimensionality reduction and sampling results of the DE-SNE algorithm, we visualize the intermediate sampling results from the CIFAR10 dataset at a 10\% keeping ratio, as shown in Fig.~\ref{fig_vis}. Original images are represented by blue dots, while the final selected samples are represented by orange dots. Regardless of the variations in the sample feature space, the small subset of samples extracted by our method adequately covers the entire feature space and captures the dataset's distribution. This capability does not depend on the inductive bias of any specific DNN, as no DNN is employed during the sampling process. Consequently, our proposed method exhibits superior generalization capability compared to the SOTA of algorithms, \ie, DQ and DQAS, and the SOTA of architectural, \ie, NeSSA.


\section{Comparison of DDR Overheads}

\begin{figure}[ht]
\centering
\includegraphics[width=0.7\columnwidth]{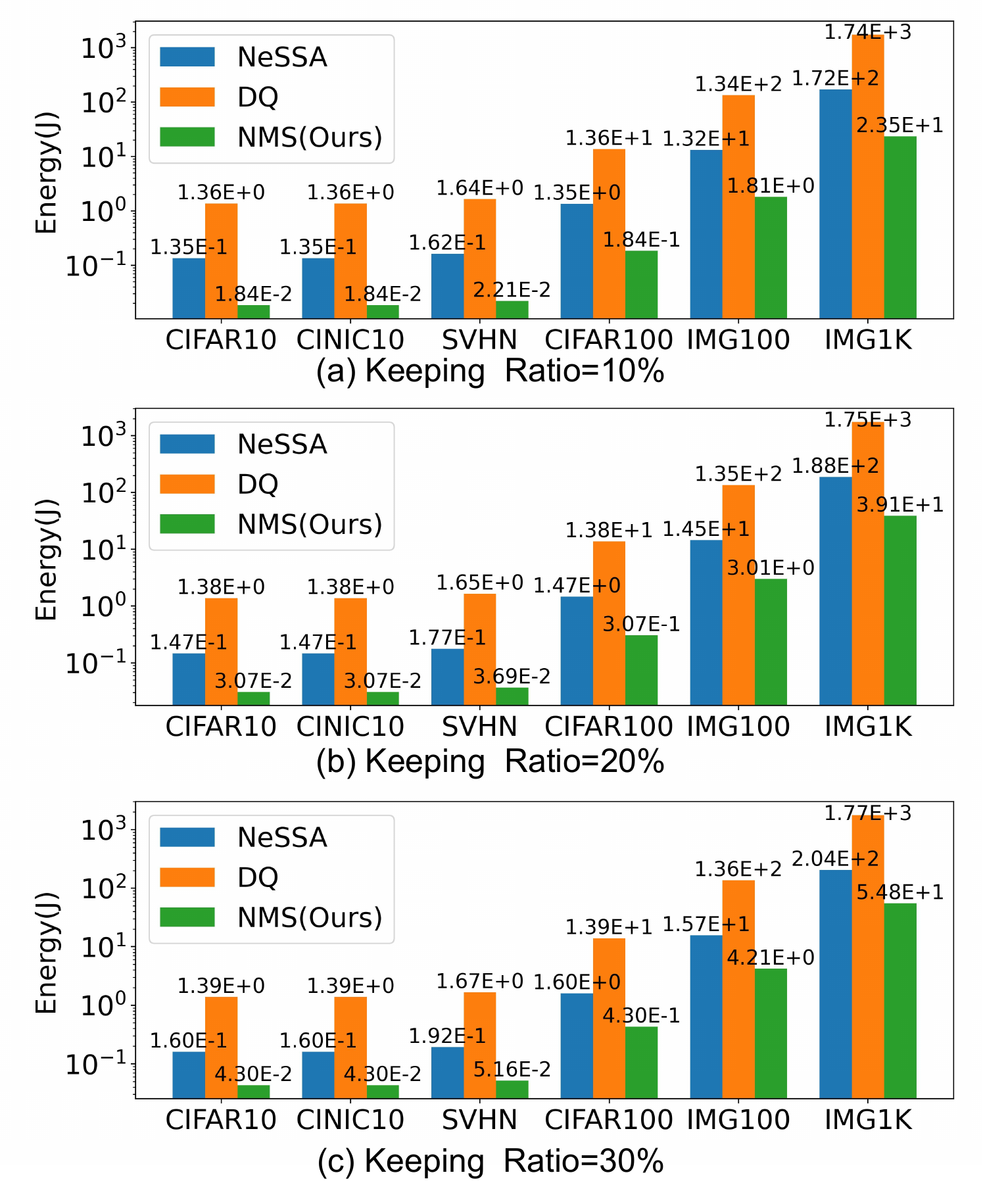}
\caption{DDR energy comparison of NeSSA, DQ and our NMS.}
\label{fig_ddr_save}
\end{figure}

To accurately assess the DDR energy consumption impact of our near-data sampling training architecture, we compare it against NeSSA and DQ across multiple datasets. As depicted in Fig.~\ref{fig_ddr_save}, our proposed Near-Memory Sampling (NMS) method achieves consistent DDR memory energy savings across various keeping ratios. 
For a keeping ratio of 10\% (Fig. ~\ref{fig_ddr_save}(a)), NMS reduces energy consumption by a factor of 7.29$\times$ to 7.34$\times$ compared to DQ, and by 73.91$\times$ to 74.04$\times$ compared to NeSSA. When the keeping ratio is set at 20\%, energy consumption decreases by a factor of 4.79$\times$ to 4.82$\times$ relative to DQ, and by 44.71$\times$ to 44.95$\times$ compared to NeSSA. At a 30\% keeping ratio, the reductions are 3.72$\times$ to 3.73$\times$ times compared to DQ, and 32.30$\times$ to 32.36$\times$ compared to NeSSA.

These results confirm that our NMS approach can substantially reduce DDR memory energy overhead compared to traditional sampling techniques. However, the data also shows that as the keeping ratio increases, the energy-saving advantage diminishes. This trend is attributed to the higher demand for long-distance PCB transmission of image samples to the training accelerator. Therefore, integrating our near-data computing architecture with an efficient DE-SNE sampling algorithm is crucial to sustaining energy savings, particularly in scenarios with lower keeping ratios.

\section{Evaluation Performance of DE-SNE}
\begin{table}[]
\centering
\caption{Ablation study of the Top-1 accuracy on $\sigma$ search for ResNet-18: t-SNE uses binary search, while DE-SNE (ours) employs differential evolution. }
\begin{tabular}{c|c|c|c|c|c}
\toprule
 Dataset           &     Method         & 10\%            & 20\%            & 30\%         &  Avg(\%)   \\ \hline
\multirow{2}{*}{CIFAR10}     & t-SNE        & 82.3          & 87.6          & 90.2         &  86.7    \\ \cline{2-6}
                             & DE-SNE & \textbf{88.0} & \textbf{90.4} & \textbf{93.5}      &  \textbf{90.6}    \\     \hline      
\multirow{2}{*}{CIFAR100}    & t-SNE        & 50.4          & 61.2          & 65.2         &  58.9    \\ \cline{2-6}
                            & DE-SNE & \textbf{54.2} & \textbf{63.7} & \textbf{68.9}       &  \textbf{62.3}    \\ \hline
\multirow{2}{*}{ImageNet-1K} & t-SNE        & 48.7          & 58.2          & 61.0         &  56.0     \\ \cline{2-6}
                            & DE-SNE & \textbf{53.6} & \textbf{62.6} & \textbf{64.5}       &  \textbf{60.2}   \\
\bottomrule
\end{tabular}
\label{vs_tsne}
\end{table}

In our ablation study, we compare the proposed DE-SNE method with the traditional t-SNE sampling approach, demonstrating the advantages of our differential evolution technique. As shown in Table \ref{vs_tsne}, DE-SNE consistently outperforms t-SNE across various datasets, indicating its superior performance in generating low-dimensional manifold representations. For the CIFAR10 dataset, DE-SNE achieves an average accuracy improvement of 6.2\% over t-SNE, with notable performance metrics of 88.0\%, 90.4\%, and 93.5\% at different keeping rates (10\%, 20\%, and 30\%, respectively). Similarly, on CIFAR100, DE-SNE yields an improvement of 3.8\% in average accuracy. Finally, for ImageNet-1K, DE-SNE again shows a significant advantage, with an average accuracy boost of 4.9\%. These results underscore the effectiveness of DE-SNE in enhancing generalization and stability, addressing the computational bottleneck associated with t-SNE's binary search process. By reducing perplexity errors, DE-SNE provides a robust alternative for low-dimensional sampling, making it particularly suitable for applications requiring efficient model training on edge devices.

\begin{figure}[ht]
\centering
\includegraphics[width=1\columnwidth]{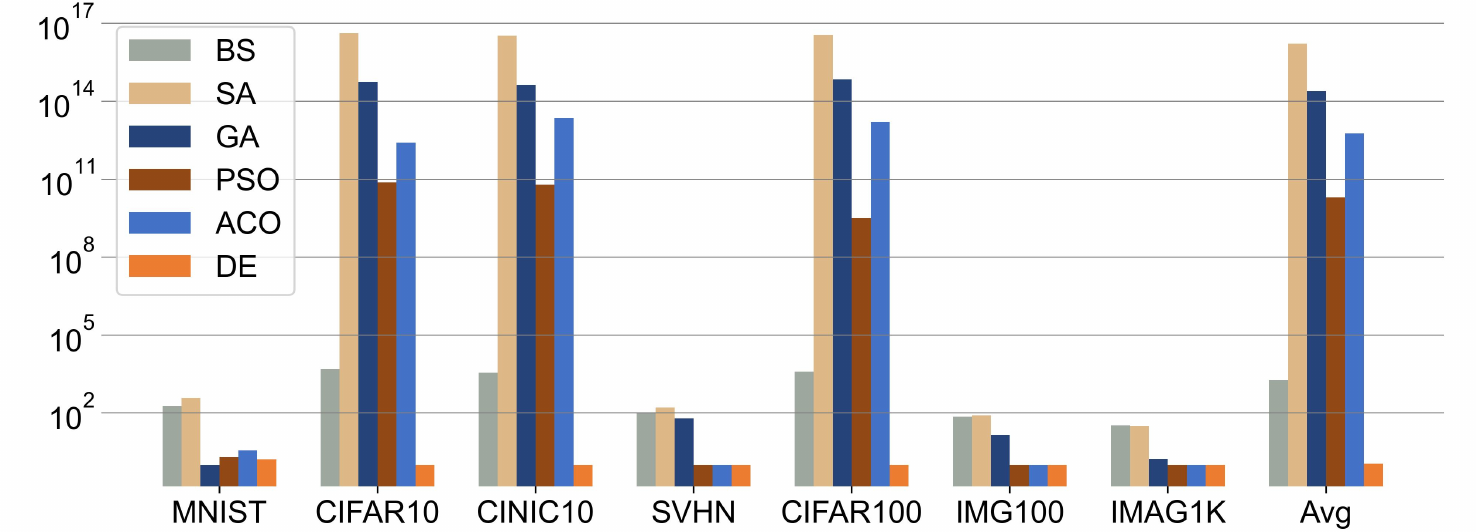} 
\caption{Perplexity error comparison of different optimization algorithms.}
\label{fig_de_error_cpu_latency} 
\end{figure}

To further investigate the reasons behind the superior accuracy achieved by DE-SNE, we compare various efficient and robust optimization algorithms with our proposed Differential Evolution (DE) algorithm. As shown in Fig.~\ref{fig_de_error_cpu_latency}, the DE algorithm achieves error reductions of 3, 16, 14, 10, and 12 orders of magnitude on six commonly used datasets, compared to binary search, simulated annealing, genetic algorithms, particle swarm optimization, and ant colony optimization, respectively. These significant accuracy improvements are attributed to our analysis and understanding of the nature of the perplexity search problem, characterized as a non-convex, non-differentiable search problem. The classic algorithms, such as Binary Search in t-SNE, and GA, PSO, and ACO, are suitable only for discrete problems and path optimization. In contrast, our proposed DE is well-suited for continuous optimization problems, demonstrating superior performance. The aforementioned experiments logically demonstrate that the superior training accuracy of DE-SNE arises from the inability of other algorithms to effectively address the perplexity issue. This further substantiates the necessity of the DE-SNE algorithm proposed in this study.


\begin{figure}[ht]
\centering
\includegraphics[width=1.0\columnwidth]{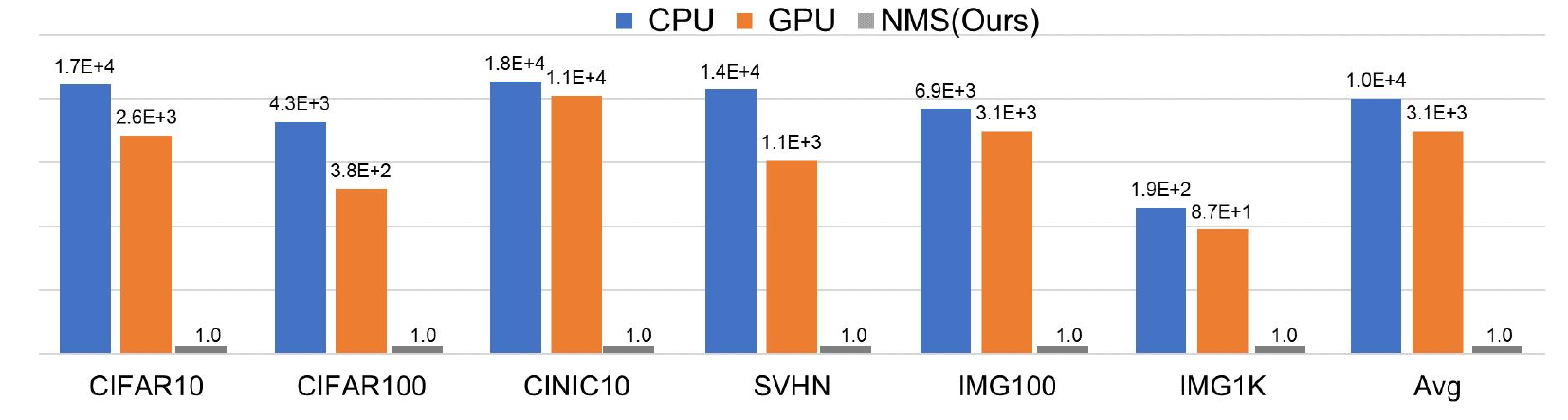}  
\caption{Normalized energy of AMD 64 Core CPU, NVIDIA 4090 GPU and proposed DE-SNE accelerator.} 
\label{fig_tsne_acc_cpu_gpu} 
\end{figure} 

\begin{figure}[ht]
\centering
\includegraphics[width=0.99\columnwidth]{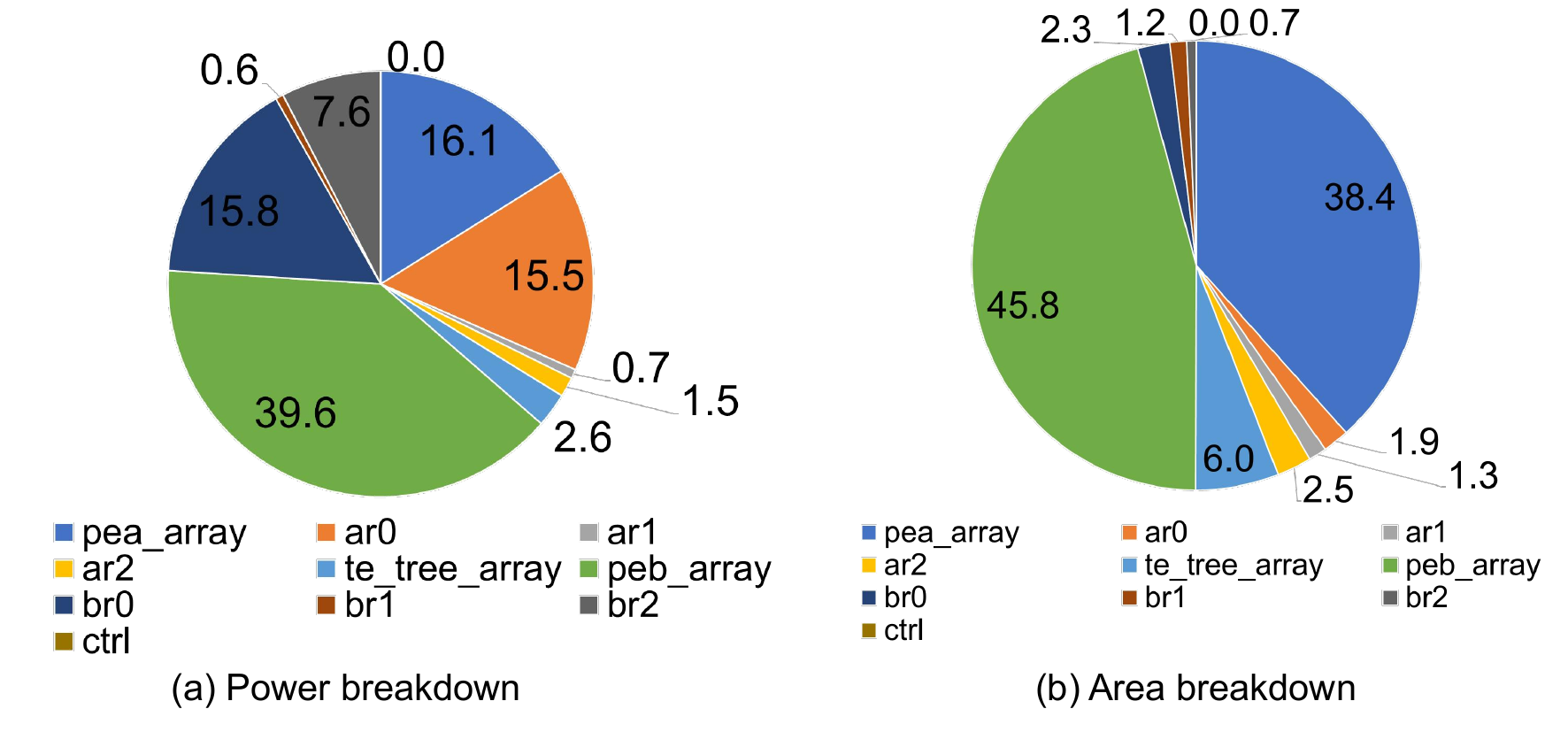}
\caption{Power and area breakdown of proposed DE-SNE accelerator.} 
\label{fig_tsne_acc_breadown}
\end{figure}  


\section{Evaluation of the DE-SNE Accelerator Circuit}
To the best of our knowledge, there is currently no dedicated ASIC accelerator for DE-SNE. To provide a fair assessment of the proposed DE-SNE accelerator, we compare its latency and energy consumption with those of high-performance CPUs and GPGPUs. It is important to note that, as highlighted in the Introduction Section, the primary challenge faced by edge-side DNN training accelerators is energy consumption. Thus, our DE-SNE accelerator aims to achieve a high energy efficiency ratio through efficient data flow and appropriately scaled hardware resources. As depicted in Fig.~\ref{fig_tsne_acc_cpu_gpu}, our solution reduces energy consumption by four and three orders of magnitude compared to the CPU and GPGPU, respectively. 

To better understand the power consumption and area overhead of the various modules within the DE-SNE accelerator, we perform a breakdown of each circuit module. As illustrated in Fig.~\ref{fig_tsne_acc_breadown}(a), in terms of power consumption, the two main processing arrays, \ie, PEA and PEB, account for the majority of power consumption, at 45.8\% and 38.4\%, respectively. The TE computation unit, containing only simple comparators and registers, accounts for merely 6\% of the power consumption. Among the SRAM modules, each cache module (AR0, AR1, BR0, BR1, and BR2) consumes less than 2\% of the total power. Regarding area, as shown in Fig.~\ref{fig_tsne_acc_breadown}(b), the processing arrays PEA and PEB occupy significant area overhead. Additionally, the largest SRAM storage units, \ie, AR0 and BR0, each occupy 15.5\% and 15.8\% of the circuit area, respectively.

 \begin{table*}[ht]

\centering
\caption{Hardware performance comparison with THETA.}	
\begin{tabular}{c|c|c|c|c|c|c|c|c}
\toprule
Metric                         & DAC'19 & DAC'20 & SparTANN & GANPU  & S.VLSI'20 & S.VLSI'21 & TVSLI'22 & Our Work \\ \hline
Process {[}nm{]}               & 45     & 65     & 65       & 65     & 65        & 28        & 28       & 28       \\ \hline
SRAM {[}KB{]}                  & -      & 364    & 105      & 676    & 338       & 643       & 388      & 436      \\ \hline
Frequency {[}MHz{]}            & 500    & -      & 250      & 200    & 200       & 440       & 200      & 200      \\ \hline
Precision                      & FXP32  & FXP16  & FP16/32  & FP8/16 & FP8/16    & FP8/16    & FP8      & FP8      \\ \hline
Throughput {[}TOPS{]}          & -      & -      & 0.38     & 24.13  & 18.01     & 58.74     & 40.12    & 66.87    \\ \hline
Speedup {[}times{]}            & 3.13   & 2.5    & 11.87    & 22.34  & 29.52     & 65.26     & 157.12   & 261.86   \\ \hline
Energy Efficiency {[}TOPS/W{]} & -      & -      & 0.648    & 68.12  & 73.87     & 148.91    & 148.24   & 176.12   \\ \hline
Energy Reduction {[}times{]}   & 2.94   & 2.9    & 6.67     & 41.04  & 51.3      & 60.04     & 82.21    & 137.01   \\ \hline
Power {[}mW{]}                 & 117    & -      & 590      & 647    & 425       & 363       & 320      & 380      \\ 
\bottomrule
\end{tabular}
\label{exp_vs_theta}
\end{table*}

\section{Overall System Evaluation}
\label{Sparse_Accelerator}

To explore the contribution of our proposed edge-side sampling training scheme to sparse training acceleration systems, we integrate the DE-SNE sampling accelerator with the sparse training accelerator proposed in THETA and collect data on power consumption, throughput, and energy efficiency. For a fair comparison, we assume a keeping ratio of 60\%, as the accuracy degradation is less than 1\% compared to training on the full dataset at this keeping ratio. As shown in TABLE~\ref{exp_vs_theta}, our designed system achieves an efficiency of 176.12 TOPS/W compared to previous sparse accelerator designs, which is 1.2 times higher than the SOTA approach. Additionally, energy consumption is reduced by 1.6 times.

\section{Conclusion}
In conclusion, this paper presents a novel approach to edge-based deep neural network (DNN) training, addressing critical challenges related to energy consumption and generalization in the context of large datasets. By introducing DE-SNE, a DNN-free sample selection algorithm inspired by the human brain's nonlinear manifold stationary, we successfully mitigate the inductive bias associated with traditional DNN-based methods. Furthermore, by implementing this algorithm within a near-memory computing architecture, we significantly reduce DDR energy consumption, thereby making edge training more practical. Our experimental results demonstrate that the proposed NMS system not only achieves superior accuracy compared to state-of-the-art methods but also offers considerable improvements in hardware efficiency. This work lays the foundation for more energy-efficient and generalizable DNN training on edge devices, marking a significant advancement in the field of edge computing.

%


\vfill

\end{document}